\documentclass[10pt,twocolumn,letterpaper]{article}

\usepackage[pagenumbers]{cvpr} 
\usepackage{graphicx} 
\usepackage{xcolor}   
\usepackage{tikz}     
\usepackage[table]{xcolor}
\usetikzlibrary{arrows.meta, positioning}

\definecolor{stage1}{RGB}{66,133,244}
\definecolor{stage2}{RGB}{234,67,53}
\definecolor{stage3}{RGB}{52,168,83}
\definecolor{darkbg}{RGB}{30,30,35}
\definecolor{brightbg}{RGB}{220,210,180}
\definecolor{nvsbg}{RGB}{180,210,230}

\usepackage[most]{tcolorbox}

\newtcolorbox{disclaimerbox}{
    colback=gray!10,     
    colframe=gray!40,    
    boxrule=0.5pt,       
    arc=4pt,             
    auto outer arc,
    boxsep=5pt,
    left=6pt,
    right=6pt,
    top=4pt,
    bottom=4pt,
    enhanced jigsaw
}

\definecolor{iccvblue}{rgb}{0.21,0.49,0.74}
\usepackage[pagebackref,breaklinks,colorlinks,allcolors=iccvblue]{hyperref}

\def\paperID{*****} 
\def\confName{CVPR}
\def\confYear{2026}

\title{NTIRE 2026 3D Restoration and Reconstruction in Real-world Adverse Conditions: RealX3D Challenge Results}

\author{
Shuhong Liu\thanks{S. Liu, C. Bao, Z. Cui, X. Chu, B. Ren, L. Gu, X. Chen, M. Li, L. Ma, M. Conde and R. Timofte were the challenge organizers, while the other authors participated in the challenge or corrupted the benchmark. Each team described its own method in the report. NTIRE 2026: \url{https://cvlai.net/ntire/2026/}. Webpage: \url{https://i2wm.github.io/3DRR_2026/}. RealX3D Benchmark: \url{https://huggingface.co/datasets/ToferFish/RealX3D}.} \quad Chenyu Bao$^{\dag}$ \quad Ziteng Cui$^{\dag}$ \quad Xuangeng Chu$^{\dag}$ \quad Bin Ren$^{\dag}$ \quad Lin Gu$^{\dag}$ \\
Xiang Chen$^{\dag}$ \quad Mingrui Li$^{\dag}$ \quad Long Ma$^{\dag}$ \quad Marcos V.\ Conde$^{\dag}$ \quad Radu Timofte$^{\dag}$ \quad Yun Liu \\
Ryo Umagami \quad Tomohiro Hashimoto \quad Zijian Hu \quad Yuan Gan \quad Tianhan Xu \quad Yusuke Kurose \\
Tatsuya Harada \quad Junwei Yuan \quad Gengjia Chang \quad Xining Ge \quad Mache You \quad Qida Cao \\ Zeliang Li \quad Xinyuan Hu \quad Hongde Gu \quad Changyue Shi \quad Jiajun Ding \quad Zhou Yu \quad Jun Yu \\ Seungsang Oh \quad Fei Wang \quad Donggun Kim \quad Zhiliang Wu \quad Seho Ahn \quad Xinye Zheng \\
Jiangxin Dong \quad Kun Li \quad Yanyan Wei \quad Weisi Lin \quad Dizhe Zhang \quad Yuchao Chen \quad Meixi Song \\\ Hanqing Wang \quad Haoran Feng \quad Lu Qi \quad Jiaao Shan \quad Yang Gu \quad Jiacheng Liu \quad Shiyu Liu \\ Kui Jiang \quad Junjun Jiang \quad Runyu Zhu \quad Sixun Dong \quad Qingxia Ye \quad Zhiqiang Zhang \\ Zhihua Xu \quad Zhiwei Wang \quad Phan The Son \quad Zhimiao Shi \quad Zixuan Guo \quad Xueming Fu \\ Lixia Han \quad Changhe Liu \quad Zhenyu Zhao \quad Manabu Tsukada \quad Zheng Zhang \quad Zihan Zhai \\ Tingting Li \quad Ziyang Zheng  \quad Yuhao Liu \quad Dingju Wang \quad Jeongbin You \quad Younghyuk Kim \\ Il-Youp Kwak \quad Mingzhe Lyu \quad Junbo Yang \quad Wenhan Yang \quad Hongsen Zhang \quad Jinqiang Cui \\ Hong Zhang \quad Haojie Guo \quad Hantang Li \quad Qiang Zhu \quad Bowen He \quad Xiandong Meng \\ Debin Zhao \quad Xiaopeng Fan \quad Wei Zhou \quad Linzhe Jiang \quad Linfeng Li \quad Louzhe Xu \quad Qi Xu \\ Hang Song \quad Chenkun Guo \quad Weizhi Nie \quad Yufei Li \quad Xingan Zhan \quad Zhanqi Shi \\ Dufeng Zhang \quad Boyuan Tian \quad Jingshuo Zeng \quad Gang He \quad Yubao Fu \quad Weijie Wang \\ Runyi Yang \quad Deheng Zhang \quad Cunchuan Huang
}

\begin{document}
\renewcommand{\thefootnote}{\fnsymbol{footnote}}
\maketitle
\thispagestyle{empty}
\pagestyle{empty}



\begin{abstract}
This paper presents a comprehensive review of the NTIRE 2026 3D Restoration and Reconstruction (3DRR) Challenge, detailing the proposed methods and results. The challenge seeks to identify robust reconstruction pipelines that are robust under real-world adverse conditions, specifically extreme low-light and smoke-degraded environments, as captured by our RealX3D benchmark. A total of 279 participants registered for the competition, of whom 33 teams submitted valid results. We thoroughly evaluate the submitted approaches against state-of-the-art baselines, revealing significant progress in 3D reconstruction under adverse conditions. Our analysis highlights shared design principles among top-performing methods and provides insights into effective strategies for handling 3D scene degradation.
\end{abstract}

\vspace{-20pt}
\section{Introduction}
\label{sec:intro}

Real-world visual data is inevitably captured under a wide range of adverse conditions that severely corrupt input image quality~\cite{chang2026training,ge2026dual,chang2026beyond,ge2026clip}. As a fundamental task underpinning applications such as autonomous driving \cite{yan2024street,chen2025dggt}, robotic navigation \cite{lisgs2025,liumg2025}, and augmented reality \cite{lidense2025,fei20243d}, 3D scene reconstruction demands high-fidelity multi-view inputs to produce accurate geometry and appearance. However, current reconstruction methods, i.e., Neural Radiance Field (NeRF) and 3D Gaussian Splatting (3DGS) \cite{mildenhall2021nerf,kerbl20233d,chen2024survey}, are predominantly developed and benchmarked under controlled, clean-capture settings. Although recent studies \cite{cui2024aleth,levy2023seathru,liuderain2025,li2025watersplatting,cui2025luminance,liu2026denoise,liu2025i2nerf,cui2026unifying} have begun to address individual corruption types, achieving faithful reconstruction remains an open challenge \cite{liu2025realx3d}, and the scarcity of real-world degraded multi-view data further hinders progress.

\begin{figure*}[!t]
    \centering
    \includegraphics[width=\textwidth]{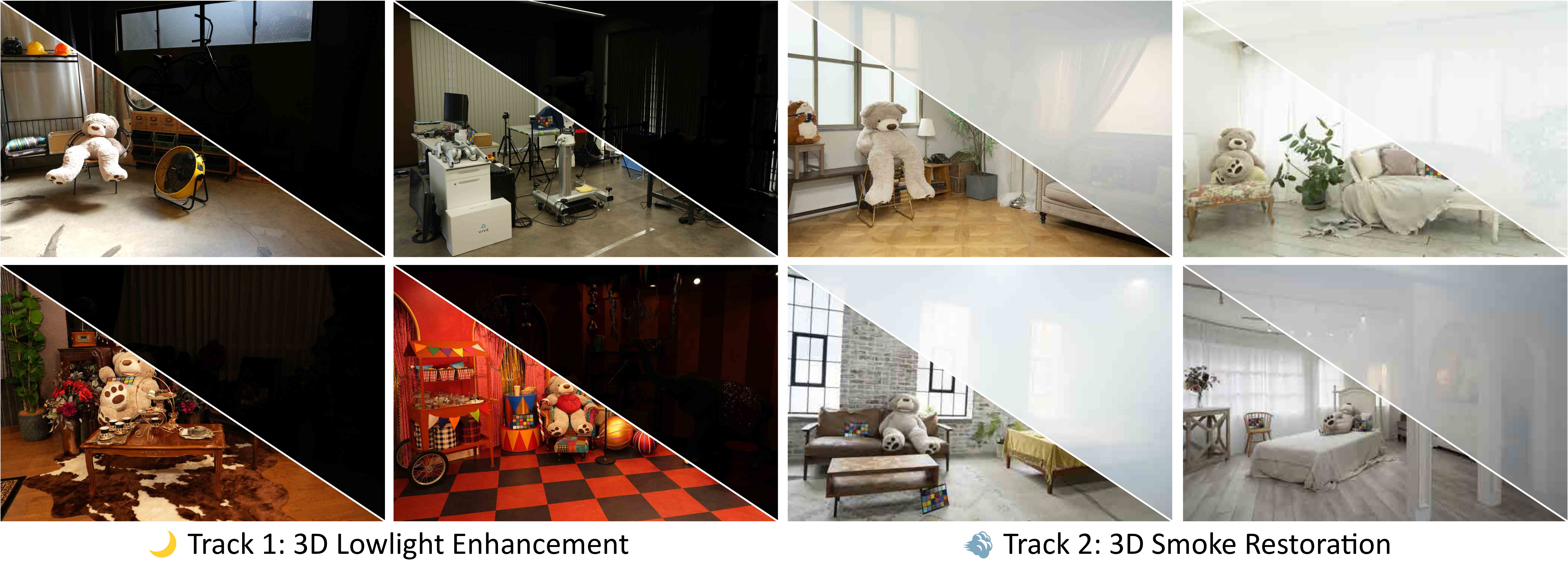}
    \caption{Visualizations of clean-degraded pairs in Track 1 \& 2. Each track features 7 distinct scenes from the RealX3D benchmark \cite{liu2025realx3d}.}
    \label{fig:teaser}
\end{figure*}

\begin{table*}[!tbp]
  \centering
  \caption{Quantitative results of the 3DRR Challenge for Track 1 and Track 2, with teams presented in descending order of rank. \colorbox{gray!20}{Gray} indicates teams that did not submit a valid factsheet and were therefore excluded from this report.}
  \label{tab:challenge_results}
  \small
  \begin{tabular}{l l c c c l l c c}
    \toprule
    \multicolumn{4}{c}{\textbf{Track 1: 3D Lowlight Enhancement}} & & \multicolumn{4}{c}{\textbf{Track 2: 3D Smoke Restoration}} \\
    \cmidrule{1-4} \cmidrule{6-9}
    \textbf{Method Name} & \textbf{Team Name} & \textbf{PSNR}$\uparrow$ & \textbf{SSIM}$\uparrow$ & & \textbf{Method Name} & \textbf{Team Name} & \textbf{PSNR}$\uparrow$ & \textbf{SSIM}$\uparrow$ \\
    \midrule
     FuME-GS & DimV             & 23.3842 & 0.8019 & &  GenSmoke-GS~\cite{cao2026gensmoke}& PLBBL           & 20.2061 & 0.7263 \\
     CISP-GS & DLMath\_Vision   & 22.7783 & 0.7763 & & Smoke-GS~\cite{zheng20263d}& XInsight Lab    & 18.6681 & 0.6909 \\
     TCIDNet-IBGS & Insta3D          & 21.6146 & 0.7122 & &  Dehaze-then-Splat~\cite{chen2026dehaze} & Hunan Duo       & 18.3816 & 0.6583 \\
     IDEAL & IDEAL & 21.5019 & 0.7283 & & MSDG & AIIA\_LAB & 17.5486 & 0.6943 \\
     NAKA-GS~\cite{zhu2026naka} & BVG              & 21.4465 & 0.7786 & & DiT-IBGS& Insta3DDD            & 16.6687 & 0.6643 \\
     GREP-GS & Wanderer         & 20.4584 & 0.7607 & & DEPHY-GS & Diouj. El       & 15.3625 & 0.6574 \\
     IC-GS & K7MQ             & 19.2141 & 0.6814 & & SmokeGS-R~\cite{fu2026smokegs}& Windrise        & 15.2174 & 0.6657 \\
    Space-GS & XInsight Lab     & 18.8954 & 0.6793 & & HAD-GS& ZZZ             & 14.8923 & 0.6103 \\
     ELoG-GS~\cite{liu2026elog} & LowLight Wizards & 18.6626 & 0.6855 & & SDG-Smoke& DLMath\_Vision  & 14.4005 & 0.6341 \\
     AdaTone-GS & SUSTech-PCL      & 17.7898 & 0.6762 & & EE-GS & EE-GS           & 14.3759 & 0.6569 \\
     SLL-GS~\cite{guo2026reliability} & 3DV-Lowlight     & 17.7584 & 0.6382 & & 3DSmokeR& Helicopter      & 13.8623 & 0.5859 \\
     GammaGS& HangFans         & 17.4242 & 0.6395 & & MonoSmokeGS& MonoSmokeGS     & 13.6479 & 0.6251 \\
     DarkIR-GS & FJNU-STAR        & 16.6617 & 0.6754 & & CPG-SGS& AAA             & 13.5287 & 0.6181 \\
     3DLLR & Stamina          & 16.5647 & 0.6914 & & DH-GS & HNU & 12.4948 & 0.5905 \\
     LLE-GS & AAA              & 16.4346 & 0.6324 & & TD-GS& FFFFYB          & 12.0852 & 0.5335 \\
     \cellcolor{gray!20}- & \cellcolor{gray!20}DUTCS & \cellcolor{gray!20}16.2584 & \cellcolor{gray!20}0.5765 & &  &  &   &  \\
     Harmony3D & RunAI & 15.4799 & 0.5676 &  &  &  &  &  \\
     AIC-GER & CC & 13.9551 & 0.5924 &  &  &  &  &  \\
    \bottomrule
  \end{tabular}
\end{table*}

To address the aforementioned challenges, we launched the first 3DRR Challenge at the 2026 NTIRE workshop. The goals of the challenge are threefold: (1) to highlight the critical problem of reconstruction under adverse conditions, (2) to offer a comprehensive evaluation benchmark, and (3) to drive research progress in this emerging field. The challenge features two tracks on two representative degradations, low-light and smoke.

This challenge is one of the NTIRE 2026 Workshop associated challenges on: ambient lighting normalization \cite{vasluianu2026ambient}, raindrop removal \cite{li2026raindrop}, smartphone ISP \cite{perevozchikov2026isp}, night-time image dehazing \cite{ancuti2026dehaze}, reflection removal \cite{cai2026reflection}, shadow removal \cite{vasluianu2026shadow}, bitstream-corrupted \cite{zou2026bitstream} and UGC \cite{li2026ugc} video restoration, AI-generated image detection \cite{gushchin2026aigc}, light-field \cite{wang2026lfsr}, efficient \cite{bin2026esr}, mobile \cite{li2026mobilesr}, infrared \cite{liu2026rsirsr}, and 3D content \cite{wang20263dsr} image super-resolution, efficient \cite{yan2026elle} and lowlight \cite{ciubotariu2026lowlight} image enhancement, image denoising \cite{sun2026denoising}, efficient \cite{aim2025efficientdeblurring} and event-based \cite{sun2026deblur} image deblurring, aberration correction \cite{sun2026aberration}, deepfake detection \cite{hopf2026deepfake}, photography retouching \cite{elezabi2026retouching}, bokeh rendering \cite{seizinger2026bokeh}, video saliency prediction \cite{moskalenko2026saliency}, RAIM \cite{qin2026raim}, video interpolation \cite{ciubotariu2026vfi}, cross-domain object detection \cite{qiu2026cdfsod}, X-AIGC quality assessment \cite{liu2026xaigc}, face restoration \cite{wang2026face} and UGC images anomaly detection \cite{zhong2026anomaly}.

\section{Tracks and Competition}
\label{sec:challenge}

\noindent\textbf{Dataset}~~
The dataset used in the 3DRR Challenge is a subset of the RealX3D dataset \cite{liu2025realx3d}, which provides diverse real-world multiview image pairs of degraded and clean captures for faithful evaluation, as shown in \Cref{fig:teaser}.\\

\noindent\textbf{Tracks}~~
The challenge consists of two tracks: (1) 3D low-light enhancement and (2) 3D smoke restoration. Each track includes seven unique scenes from the RealX3D dataset \cite{liu2025realx3d}. For each scene, degraded views and ground-truth camera poses are provided as input. Participants are required to reconstruct the 3D scenes and submit rendered novel-view synthesis (NVS) results, which are evaluated against ground-truth (GT) clean captures. During the development phase, we fully released 1 validation scene with paired degraded and clean views for training purposes, along with 4 additional development scenes containing only degraded views. During the testing phase, we released 3 scenes with degraded views only. Each scene contains \~30 images for training and \~5 images for NVS evaluation.\\

\noindent\textbf{Ranking criteria}~~
We evaluate the NVS results using PSNR and SSIM. To simplify the evaluation protocol, the final ranking is determined by the average PSNR across all scenes. SSIM is used only as a tiebreaker. Evaluation is conducted on the Codalab server. Participants are required to submit both their code and final results, which are checked by the organizers to ensure reproducibility. The challenge results of both tracks are summarized in \Cref{tab:challenge_results}.


\section{Track 1: 3D Lowlight Enhancement}
\label{subsec:t1_methods}

\subsection{FuME-GS: Fusion-Guided Multi-Stage Enhancement for 3DGS}
\label{subsec:DimV}

\begin{center}


\noindent\emph{Mache You$^{1}$, Zeliang Li$^{1}$, Hongde Gu$^{1}$, Jiangxin Dong$^{1}$}


\noindent\emph{$^{1}$Nanjing University of Science and Technology}

\end{center}

\begin{figure}[h]
    \centering
    \vspace{-1em}
    \includegraphics[width=\linewidth]{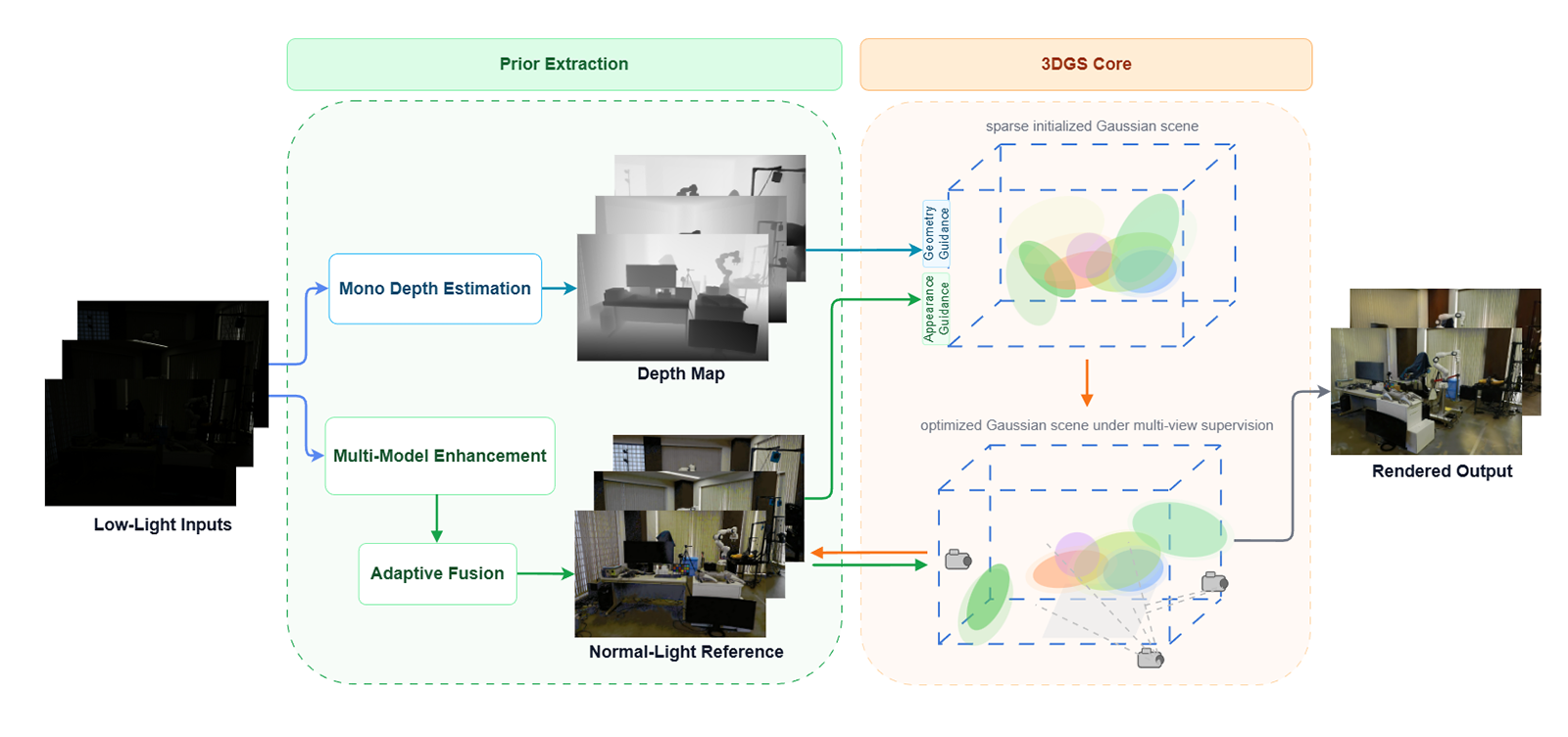}
    \caption{Overview of the proposed Fusion-Guided Multi-Stage Enhancement for 3DGS.}
    \label{fig:dimv_pipeline}
\end{figure}


\noindent\textbf{Method}~~
The authors propose a multi-stage pipeline integrating multi-model enhancement, adaptive fusion, and 3DGS. As illustrated in Fig.~\ref{fig:dimv_pipeline}, input low-light images are first processed by multiple enhancement models (Retinexformer \cite{cai2023retinexformer}, Zero-DCE \cite{guo2020zero}, ReDDiT \cite{lan2025efficient}, and HVI-CIDNet \cite{hvi}) to generate a set of restored candidates. A region fusion strategy then combines these candidates to produce a high-quality enhanced image with improved illumination, reduced noise, and preserved structural details. For novel view synthesis (NVS), a 3DGS model is optimized using a fusion-guided point cloud initialization. Specifically, depth estimated from the low-light inputs provides reliable geometric structural cues, while the enhanced images supply clean appearance priors for radiance modeling. 



\subsection{CISP-GS: Multi-Branch GS with ISP Supervision and Frequency-Split Target Fusion}
\label{subsec:dlmath_vision}

\begin{center}


\noindent\emph{Seungsang Oh$^{1}$, Donggun Kim$^{1}$, Seho Ahn$^{1}$}


\noindent\emph{$^{1}$Korea University}

\end{center}

\vspace{-1em}

\begin{figure}[h]
    \centering
    \includegraphics[width=\linewidth]{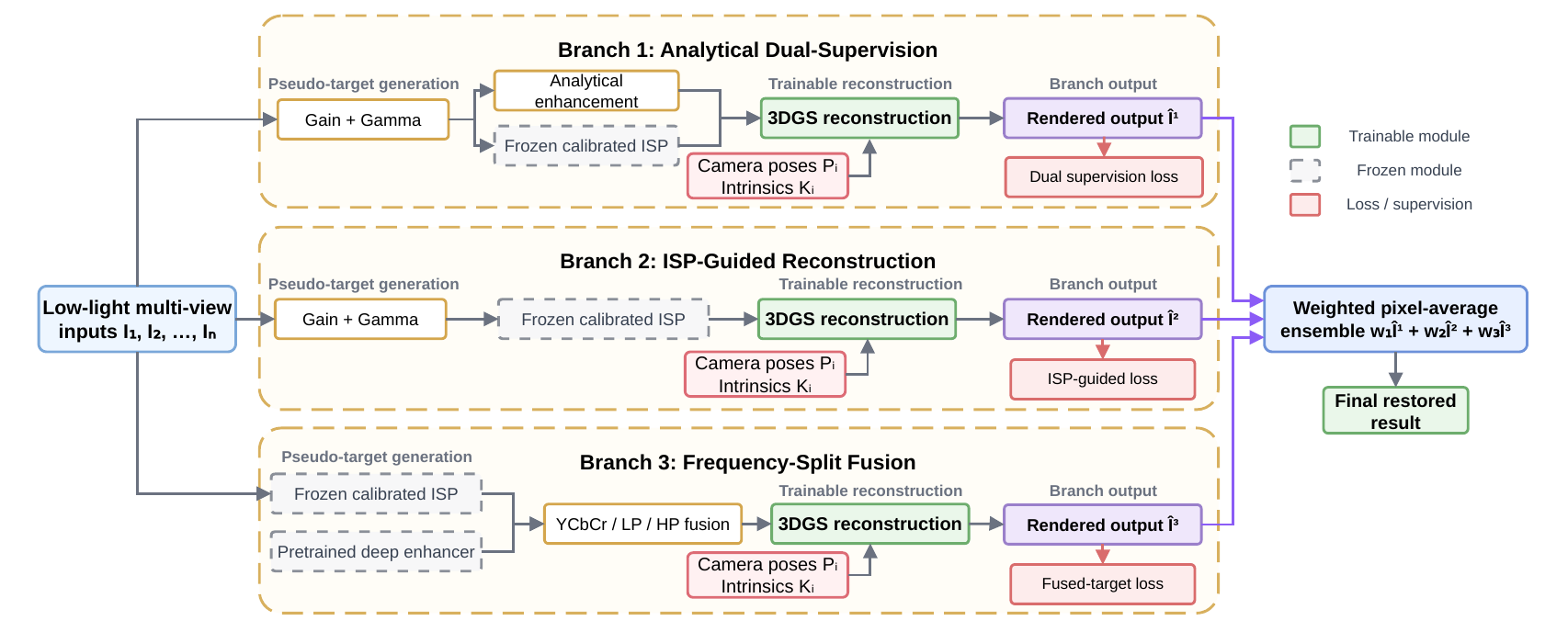}
    \caption{Overall architecture of the proposed multi-branch low-light 3DGS framework.}
    \label{fig:dlmath_arch}
\end{figure}


\noindent\textbf{Method}~~
The authors propose 3DGS4LL, a multi-branch 3DGS framework (see Fig.~\ref{fig:dlmath_arch}) that optimizes three complementary branches with distinct pseudo GT supervision mechanisms. Branch 1 utilizes analytical dual-supervision, combining an analytically enhanced target (gain, gamma, white-balance, denoising) with a calibrated ISP target for a stable, camera-consistent tone response. Branch 2 adopts a deterministic ISP-guided strategy, optimizing directly toward a single PGT from a frozen calibrated ISP to anchor global scene appearance. Branch 3 employs frequency-split fusion in the YCbCr domain, preserving stable low-frequency structure and chroma from the ISP while injecting high-frequency luminance details from a pretrained deep enhancement network. Finally, a weighted pixel-average ensemble of the three rendered outputs is applied during inference to robustly balance structural consistency, global tone stability, and local detail recovery.\\




\noindent\textbf{Training Details}~~
The 3DGS backbone is trained from scratch. Densification is enabled during the early and middle stages to improve geometric coverage, followed by later-stage refinement. The SH degree is progressively increased for stable color modeling. Both the calibrated ISP model and the pretrained Retinex-based network (used in Branch 3) remain frozen throughout the optimization process, serving solely for PGT generation.

\subsection{TCIDNet-IBGS}
\label{subsec:insta3d}

\begin{center}


\noindent\emph{Dizhe Zhang$^{1}$, Meixi Song$^{1}$, Haoran Feng$^{1}$, Lu Qi$^{1}$}


\noindent\emph{$^{1}$Insta360 research}

\end{center}


\begin{figure}[!h]
\centering
\includegraphics[width=\linewidth]{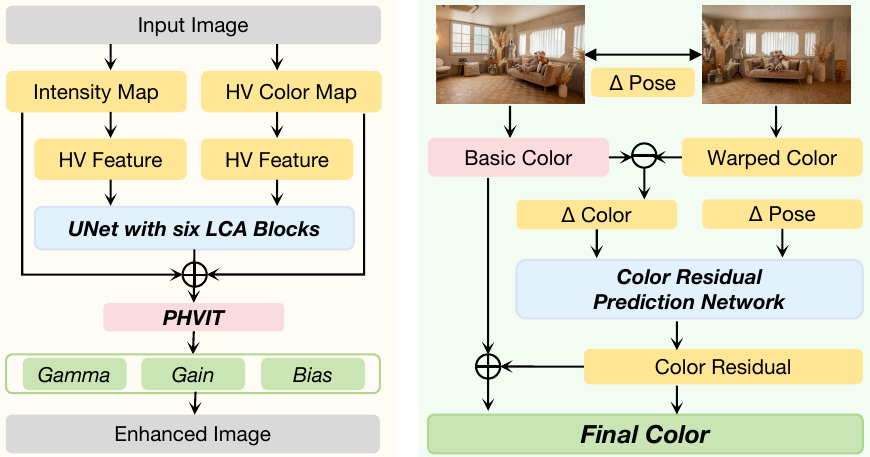}
\caption{Overview of the proposed TCIDNet-IBGS.}
\label{fig:tcidnet_ibgs}
\vspace{-0.5em}
\end{figure}

\noindent\textbf{Method}~~
The authors propose TCIDNet-IBGS, a unified framework combining illumination-aware image restoration with geometry-driven rendering (see Fig.~\ref{fig:tcidnet_ibgs}). For restoration, CIDNet \cite{hvi} is adopted as the backbone, mapping RGB inputs into the HVI color space to process intensity and chromatic information via two coordinated streams. Multi-scale cross-stream interactions are facilitated by lightweight attention-guided blocks, jointly optimizing brightness recovery and color correction. A subsequent compact global tone-correction head predicts image-adaptive gamma, channel gain, and channel bias, ensuring stable scene-level photometric calibration while maintaining a lightweight architecture. The enhanced views are subsequently fed into IBGS, an extended 3DGS module incorporating image-guided geometric and photometric constraints. Scene initialization constructs per-view neighboring source lists based on camera distance, view angle, and pose differences to provide geometry-relevant multi-view support. The 3D Gaussians learn spatial, appearance, normal, and offset attributes.

\subsection{IDEAL: Illumination Disentanglement and Enhancement for 3D Low-Light Reconstruction}
\label{subsec:ideal}

\begin{center}

\noindent\emph{Jiaao Shan$^{1}$ \quad $^{1}$Harbin Institute of Technology}
\end{center}

\begin{figure}[!h]
\centering
\includegraphics[width=\linewidth]{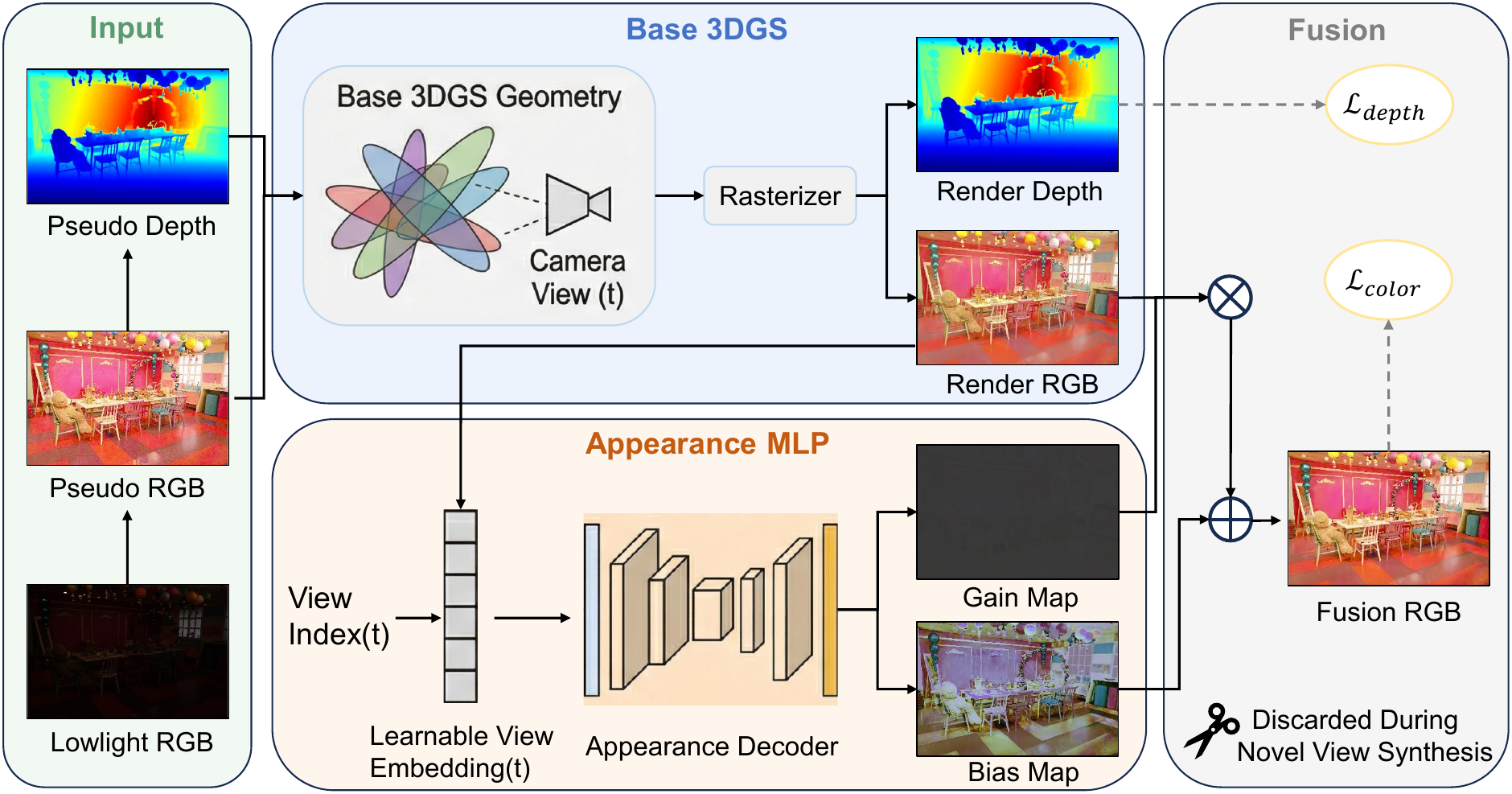} 
\caption{Overall architecture of IDEAL. The Appearance MLP predicts Gain and Bias maps to absorb view-dependent artifacts from Pseudo RGB during training, supervised by $\mathcal{L}_{color}$ and $\mathcal{L}_{depth}$. At inference, the Appearance MLP is discarded.}
\vspace{-1em}
\label{fig:ideal}
\end{figure}


\noindent\textbf{Method}~~
The authors propose IDEAL shown in Fig. \ref{fig:ideal}. First, a fine-tuned Zero-DCE++ \cite{Zero-DCE++} generates normal-illumination pseudo-GT images, while Depth-AnythingV3 \cite{depthanything3} extracts pseudo-depth maps. To prevent baking 2D artifacts into the 3D geometry, IDEAL disentangles physical geometry from noisy illumination priors using a base 3DGS \cite{kerbl20233d} and a lightweight Appearance Decoder (MLP). The MLP utilizes view-dependent learnable embeddings to absorb noise and illumination inconsistencies, effectively shielding the base 3DGS. For NVS, the MLP is discarded. The overall loss function is formulated as:
\( \mathcal{L} = \mathcal{L}_{color} + \lambda_{d} \mathcal{L}_{depth} + \lambda_{e} \mathcal{L}_{emb} + \lambda_{a} \mathcal{L}_{app} + \lambda_{n} \mathcal{L}_{near} \)
where $\mathcal{L}_{color}$ combines $L_1$ and SSIM, $\mathcal{L}_{depth}$ is a Pearson depth loss, $\mathcal{L}_{emb}$ and $\mathcal{L}_{app}$ regularize view embeddings and MLP outputs, and $\mathcal{L}_{near}$ clears near-camera floaters.\\

\noindent\textbf{Training Details}~~
Implemented in PyTorch and optimized with Adam.
The loss weights are set to $\lambda_{d}=0.02$, $\lambda_{e}=0.01$, $\lambda_{a}=0.1$, and $\lambda_{n}=0.1$. The 3DGS position gradient threshold is increased to 0.0004 to prevent noise-induced over-densification and accelerate training. All other hyperparameters follow standard configurations.

\subsection{NAKA-GS: A Bionics-inspired Dual-Branch Naka Correlation and Progressive Point Pruning for Low-Light 3DGS}
\label{subsec:BVG}

\begin{center}


\noindent\emph{Runyu Zhu$^{1}$, Sixun Dong$^{1}$, Qingxia Ye$^{1}$, Zhiqiang Zhang$^{1}$\\ Zhihua Xu$^{1}$}


\noindent\emph{$^{1}$China University of Mining and Technology (Beijing)}

\end{center}

\vspace{-0.5em}
\begin{figure}[!h]
    \centering
    \includegraphics[width=1\linewidth, trim=0cm 1cm 0cm 0cm, clip]{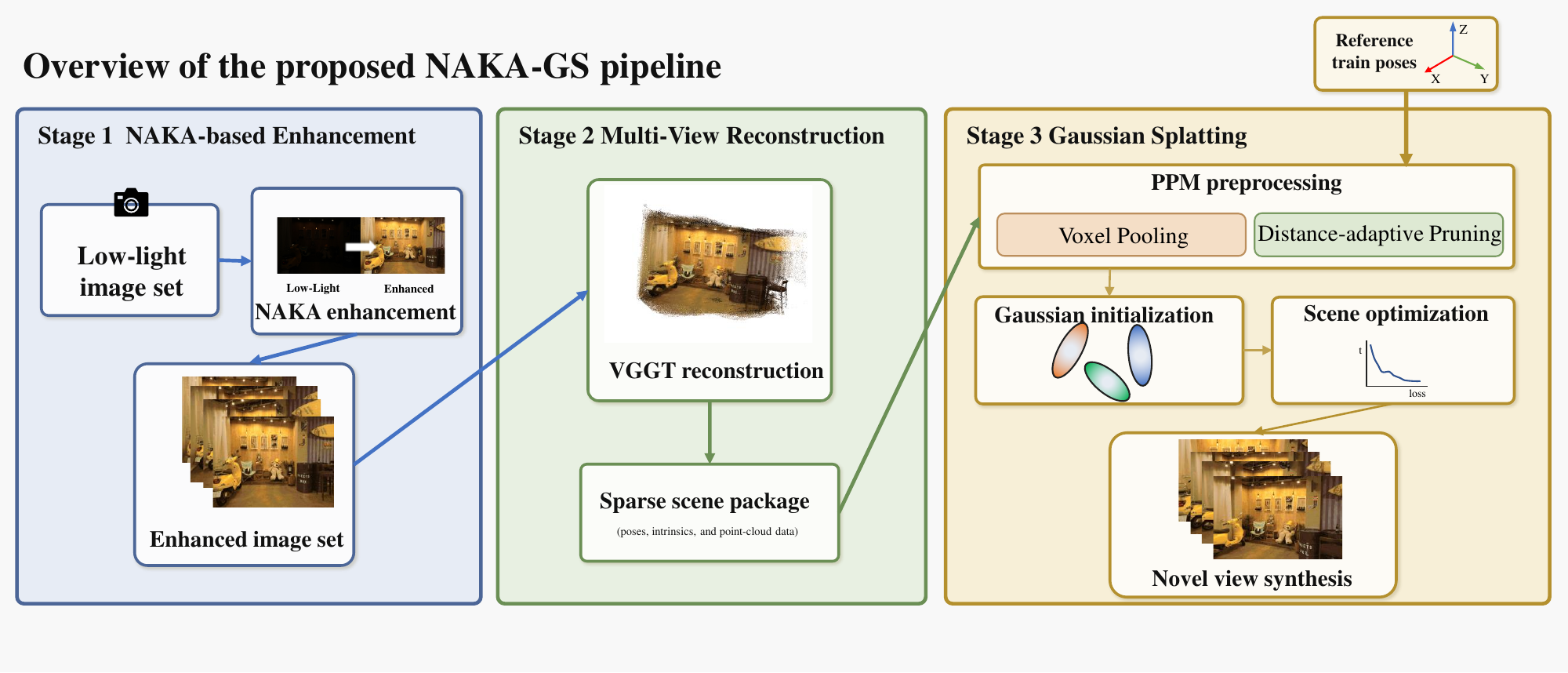}
    \caption{Overview of the proposed NAKA-GS~\cite{zhu2026naka} pipeline, including NAKA-based enhancement, VGGT-based multi-view reconstruction, and Gaussian Splatting with PPM preprocessing.}
    \label{fig:nakags_pipeline}
\end{figure}


\noindent\textbf{Method}~~
The authors propose NAKA-GS~\cite{zhu2026naka}, a pipeline integrating a Naka-guided chroma-correction model and a Point-cloud Pruning Module (PPM) for low-light 3DGS (see Fig.~\ref{fig:nakags_pipeline}). To address color distortions and edge errors, the chroma-correction model combines a physics-prior Naka transform with a U-Net-style network. It processes an 18-channel representation encoding the original degradation, the Naka-enhanced result, and their residuals in both raw and normalized forms. The network decomposes the Naka image into low- and high-frequency components, applying multiplicative-additive corrections exclusively to the low-frequency part while preserving structural details by directly injecting the high-frequency residuals. The enhanced images are then processed by a pre-trained VGGT model \cite{wang2025vggt} to generate dense point clouds. Prior to 3DGS initialization, the PPM refines these dense priors through coordinate alignment, voxel pooling, and distance-adaptive progressive random pruning.\\


\noindent\textbf{Training Details}~~
The chroma-correction model is trained on 175 images from the LOM dataset \cite{cui2024aleth} and RealX3D validation set \cite{liu2025realx3d}. The model is optimized using AdamW with an initial learning rate of $2\times10^{-4}$ for 200 epochs, a batch size of 8, and $256\times256$ random crops. The compound loss function combines a base loss (RGB reconstruction, YCbCr chroma consistency, SSIM, Sobel edge, VGG perceptual, and regularization) with a Gray Edge Mask Loss and a Bright Mask Loss to emphasize structurally informative and relatively bright regions, respectively. For the 3DGS backend, the PPM applies a voxel pooling size of 0.01 and 6 pruning iterations. The subsequent 3DGS optimization runs for 8,000 steps with a batch size of 1.

\subsection{GREP-GS: Global-Residual Decomposition with Hybrid Edge Prior for Low-Light 3DGS}
\label{subsec:wanderer}

\begin{center}


\noindent\emph{Zhiwei Wang$^{1}$, Zhimiao Shi$^{1}$}


\noindent\emph{$^{1}$National University of Defense Technology}

\end{center}

\vspace{-0.5em}
\begin{figure}[!h]
    \centering
    \includegraphics[width=\linewidth]{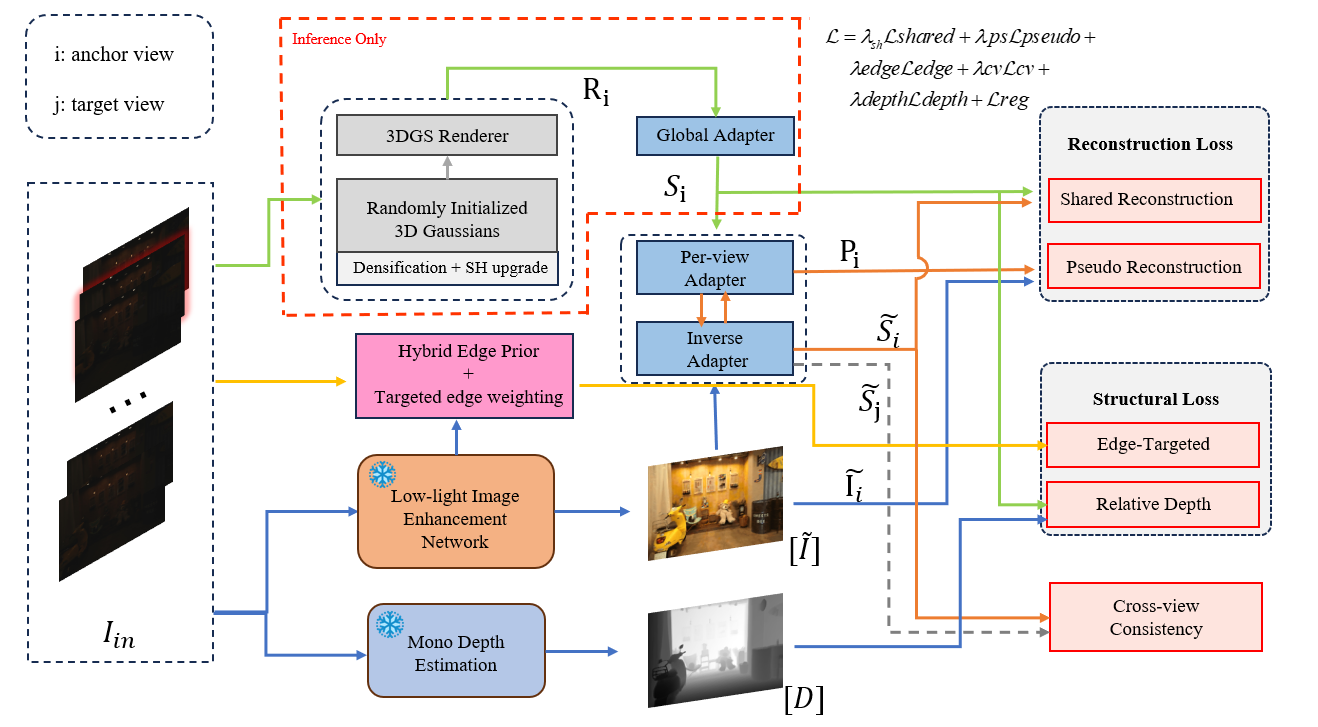}
    \caption{Overview of GREP-GS. Pseudo-enhanced targets and relative depth priors are first extracted from low-light inputs, then 3DGS is optimized in a canonical display space using global and per-view adapters to handle view-specific enhancement bias.}
    \label{fig:wanderer}
\end{figure}


\noindent\textbf{Method}~~
The proposed method leverages pretrained networks as priors to mitigate cross-view inconsistency, texture-detail loss, and noise during reconstruction as shown in Fig.~\ref{fig:wanderer}. Specifically, Retinexformer \cite{cai2023retinexformer} and MoGe2 \cite{wang2025moge} are employed to generate enhanced images and relative depth maps from the original low-light inputs. The key idea is to separate the learning target into a shared canonical display space and a view-specific residual correction space: the canonical display space captures geometry-consistent scene appearance for novel-view synthesis, while the residual branch explains the view-dependent deviations introduced by single-image enhancement.
The pipeline starts from a randomly initialized set of 100K Gaussians and optimizes geometry, opacity, anisotropic scale, and SH coefficients in a standard 3DGS renderer. For a training view \(i\) with camera pose \(T_i\), the 3DGS renderer predicts RGB, opacity, and expected depth as \((R_i,A_i,\hat{D}_i)=\mathcal{G}_{\Theta}(T_i)\). The rendered output is first transformed by a global photometric adapter into the canonical display space:
\(
    S_i(x)=\operatorname{clip}\!\left(\mathbf{M}_gR_i(x)+\mathbf{b}_g,0,1\right),
\)
where \(\mathbf{M}_g=\mathbf{I}+\Delta\mathbf{M}_g\). A lightweight residual adapter then transforms the canonical output to the pseudo target space of each training image:
\(
    P_i(x)=\operatorname{clip}\!\left(\mathbf{s}_i\odot S_i(x)+\mathbf{t}_i,0,1\right).
\)
The shared-space supervision target is obtained by inverse residual adaptation.



\subsection{IC-GS: Illumination-Calibrated 2DGS}
\label{subsec:k7mq}

\begin{center}

\noindent\emph{Zixuan Guo$^{1}$, Changhe Liu$^{1}$, Manabu Tsukada$^{1}$}


\noindent\emph{$^{1}$The University of Tokyo}

\end{center}

\vspace{-1.5em}

\begin{figure}[h]
\centering
\includegraphics[width=\linewidth]{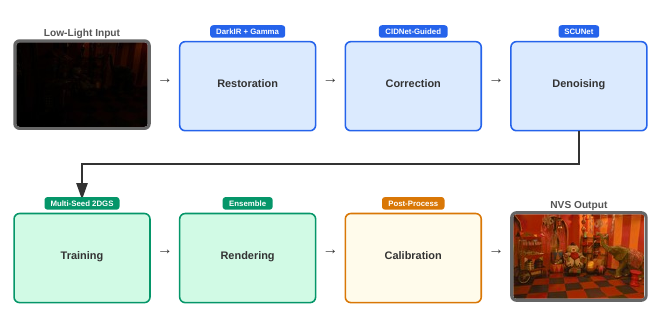}
\caption{Overview of IC-GS. Top: low-light enhancement via DarkIR with gamma boost, CIDNet-guided color correction, and SCUNet denoising. Bottom: multi-seed 2DGS training with ensemble rendering and gamma+std post-processing.}
\label{fig:k7mq_pipeline}
\end{figure}


\noindent\textbf{Method}~~
They use a multi-step low-light enhancement pipeline followed by 2DGS \cite{huang20242d} for NVS (see Fig.~\ref{fig:k7mq_pipeline}). First, DarkIR \cite{darkir} restores structural details from low-light inputs, followed by a fixed gamma boost to raise overall brightness. Then, we run CIDNet \cite{hvi} on the DarkIR output to obtain per-scene brightness and color statistics as reference targets. They compute an adaptive per-channel gamma correction that maps DarkIR's brightness to CIDNet's brightness level. SCUNet \cite{zhang2023practical} then denoises the gamma-corrected images to suppress amplified noise. For NVS, we train multi-seed 2DGS models on the enhanced views, render at an earlier checkpoint, and ensemble via pixel-wise averaging to suppress floater artifacts. Finally, a per-channel gamma and standard-deviation correction, calibrated on the validation scene, is applied as post-processing to align the color distribution of the rendered outputs.\\

\noindent\textbf{Training Details}~~
All scenes share unified 2DGS hyperparameters: densification only during the early training phase, opacity culling at 0.1, combined L1 + D-SSIM loss ($\lambda_{\text{dssim}}=0.2$) and distortion regularization ($\lambda_{\text{dist}}=0.01$) to suppress scattered Gaussians. The rendered outputs per test view are averaged pixel-wise, followed by per-channel gamma and standard-deviation post-processing calibrated on the validation scene.

\subsection{Space-GS}
\label{subsec:xinsight_lab_t1}

\begin{center}


\noindent\emph{Fei Wang$^{1,2}$, Xinye Zheng$^{1}$, Zheng Zhang$^{1}$, Zhiliang Wu$^{3}$\\ Kun Li$^{4}$, Weisi Lin$^{3}$}


\noindent\emph{$^{1}$Hefei University of Technology, $^{2}$Hefei Comprehensive National Science Center, $^{3}$Nanyang Technological University, $^{4}$United Arab Emirates University}

\end{center}

\begin{figure}[h]
\centering
\includegraphics[width=\linewidth]{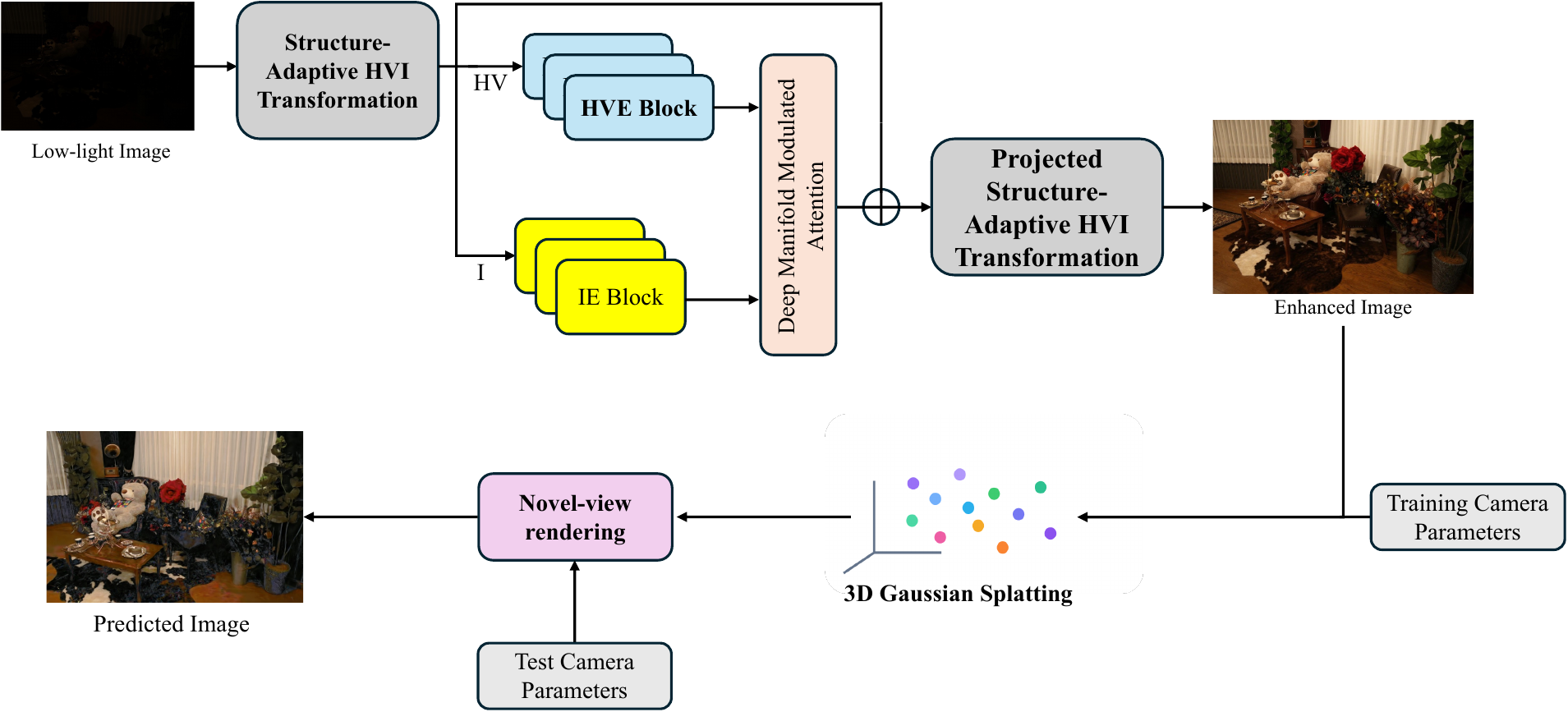}
\caption{Overview of Space-GS. Low-light inputs are split into chrominance (HV) and intensity (I) streams by SAHT, enhanced through dual-stream processing with geometry-guided DMMA interaction, and reconstructed via inverse PSAHT.}
\label{fig:xinsight_t1}
\end{figure}


\noindent\textbf{Method}~~
The authors propose Space-GS, combining GeoLight with 3DGS to preserve geometric structures and multi-view consistency. As shown in Fig.~\ref{fig:xinsight_t1}, GeoLight uses a Structure-Adaptive HVI Transformation (SAHT) module to map RGB inputs to HVI space, decoupling illumination from hue and saturation. A dual-stream backbone extracts features from chrominance and luminance streams, exchanging information via a deep manifold modulated attention (DMMA) block that maintains structural boundaries using spatial priors. A cross-attention module then aligns color and luminance distributions. Enhanced features are merged via skip connections and projected back to RGB. Finally, restored images and poses are fed into COLMAP for sparse point cloud extraction, optimizing the 3DGS representation for novel view synthesis.\\


\noindent\textbf{Training Details}~~
Models are trained on \(256 \times 256\) patches, augmented with flipping and random Gamma perturbations. The Adam optimizer is used with an initial learning rate of \(1 \times 10^{-4}\), decaying to \(1 \times 10^{-7}\) via cosine annealing. The loss function constrains both RGB and HVI domains:
\(
    \mathcal{L} = \mathcal{L}_{RGB} + \lambda_{HVI} \mathcal{L}_{HVI} + \lambda_{F} \mathcal{L}_{Fourier}
\)
where \(\mathcal{L}_{RGB}\) combines \(L_1\), SSIM, edge, and perceptual losses; \(\mathcal{L}_{HVI}\) applies \(L_1\) supervision; and \(\mathcal{L}_{Fourier}\) maintains frequency consistency.


\subsection{ELoG-GS: Extreme Low-light Optimized Gaussian Splatting}
\label{subsec:LowLightWizards}

\begin{center}


\noindent\emph{Ziyang Zheng$^{1}$, Yuhao Liu$^{1}$, Dingju Wang$^{1}$}


\noindent\emph{$^{1}$Shanghai Jiao Tong University}

\end{center}

\vspace{-1em}

\begin{figure}[h]
    \centering
    \includegraphics[width=\linewidth]{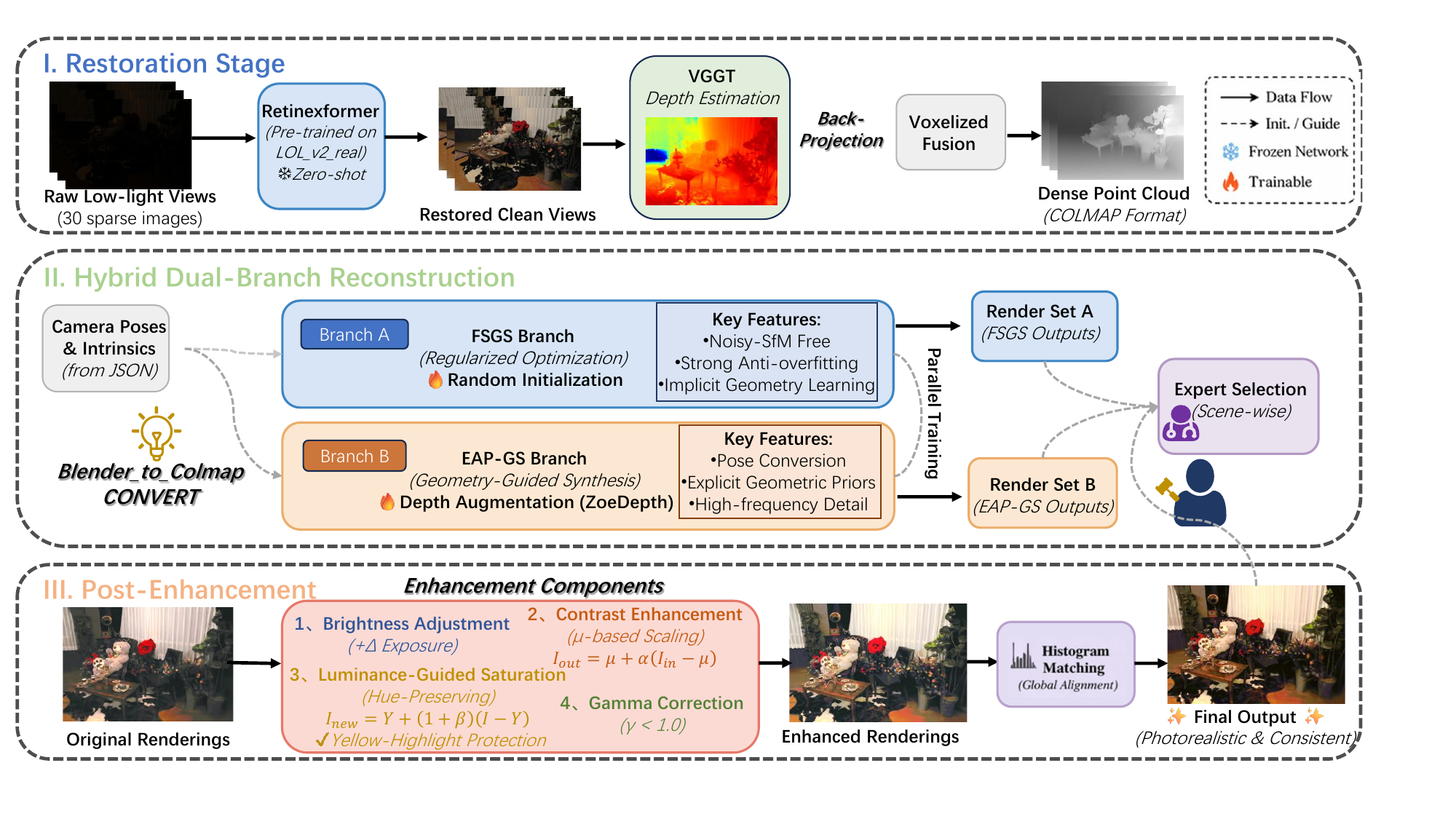}
    \caption{The overall architecture of ELoG-GS~\cite{liu2026elog}, featuring hybrid dual-branch reconstruction and post-enhancement stages.}
    \label{fig:llw}
\end{figure}


\noindent\textbf{Method}~~
The authors propose Extreme Low-light Optimized Gaussian Splatting (ELoG-GS)~\cite{liu2026elog}, an explicit ``restoration-then-reconstruction'' framework. As shown in Fig.~\ref{fig:llw}, the pipeline comprises three stages: restoration, hybrid dual-branch reconstruction, and post-enhancement. First, a zero-shot Retinexformer~\cite{cai2023retinexformer} recovers latent scene information, while VGGT~\cite{wang2025vggt}-based depth estimation and voxelized fusion provide robust geometric initialization, bypassing unreliable SfM results. Next, a dual-branch reconstruction engine integrates FSGS~\cite{zhu2024fsgs} and EAP-GS~\cite{dai2025eap}. The FSGS branch employs random initialization and regularized optimization for global stability, whereas the EAP-GS branch uses monocular depth priors to recover high-frequency textures and sharp boundaries. Finally, the better-performing branch is selected, and its rendered outputs undergo luminance-guided enhancement and Histogram Matching to align pixel distributions with scene statistics.

\subsection{AdaTone-GS}
\label{subsec:sustech_pcl}

\begin{center}


\noindent\emph{Mingzhe Lyu$^{1,2}$, Wenhan Yang$^{2}$, Jinqiang Cui$^{2}$\\ Hong Zhang$^{1}$}

\noindent\emph{$^{1}$Southern University of Science and Technology, $^{2}$Pengcheng Laboratory}

\end{center}

\begin{figure}[h]
\centering
\includegraphics[width=\linewidth]{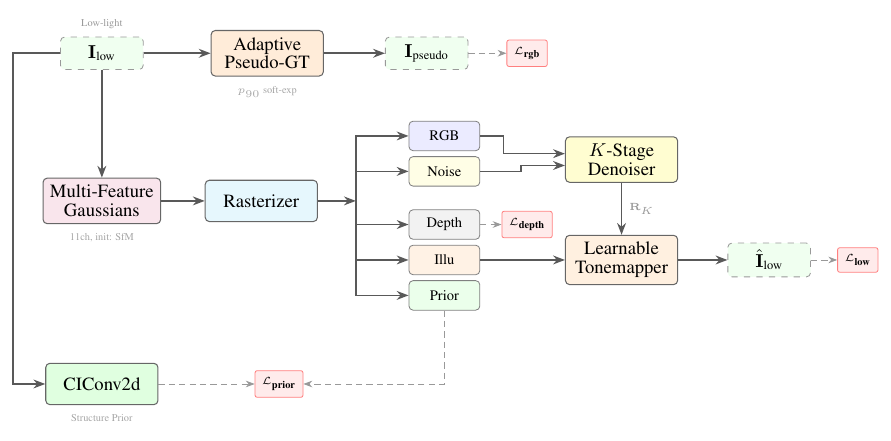}
\caption{Overview of AdaTone-GS. Adaptive pseudo-GTs supervise multi-feature Gaussians rasterized into RGB, depth, prior, noise, and illumination channels, then reconstructed via a cascaded denoiser and learnable tonemapper.}
\label{fig:sustech}
\end{figure}


\noindent\textbf{Method}~~
The authors propose AdaTone-GS. As shown in Fig.~\ref{fig:sustech}, it introduces a scene-adaptive pseudo GT generation strategy, eliminating the need for external enhancement networks. For a low-light input \(\mathbf{I}_{\text{low}}\), the 90th luminance percentile \(p_{90}\) is computed to apply a soft-exponential tone curve: \(\mathbf{I}_{\text{pseudo}} = 1 - \exp(-g \cdot \mathbf{I}_{\text{low}})\), where \(g = -\ln(0.2)/p_{90}\) adapts automatically to the scene's exposure level. Each Gaussian is augmented with depth, structure prior, noise, and illumination features (11 channels total). A cascaded \(K\)-stage denoiser progressively cleans the rendered image, while a learnable tonemapper reconstructs the low-light observation via \(\hat{\mathbf{I}}_{\text{low}} = \text{ToneMap}(\mathbf{L} \odot \mathbf{R}_K)\). Additionally, Color-Invariant Convolutions (CIConv2d) provide illumination-invariant structure priors, and depth-based reprojection enforces cross-view geometric consistency.\\


\noindent\textbf{Training Details}~~
The model is trained for 5K--10K iterations using the Adam optimizer with standard 3DGS densification.
The total loss combines \(\mathcal{L}_1\) and D-SSIM on the pseudo-GT, low-light reconstruction loss, Pearson depth loss, structure prior loss, cross-view consistency loss, and total variation (TV) regularization on the denoiser residual.

\subsection{SLL-GS: Staged Low-Light 3DGS}
\label{subsec:3dv_lowlight}

\begin{center}


\noindent\emph{Haojie Guo$^{1}$}


\noindent\emph{$^{1}$Huazhong University of Science and Technology}

\end{center}

\begin{figure}[h]
\centering
\includegraphics[width=\linewidth]{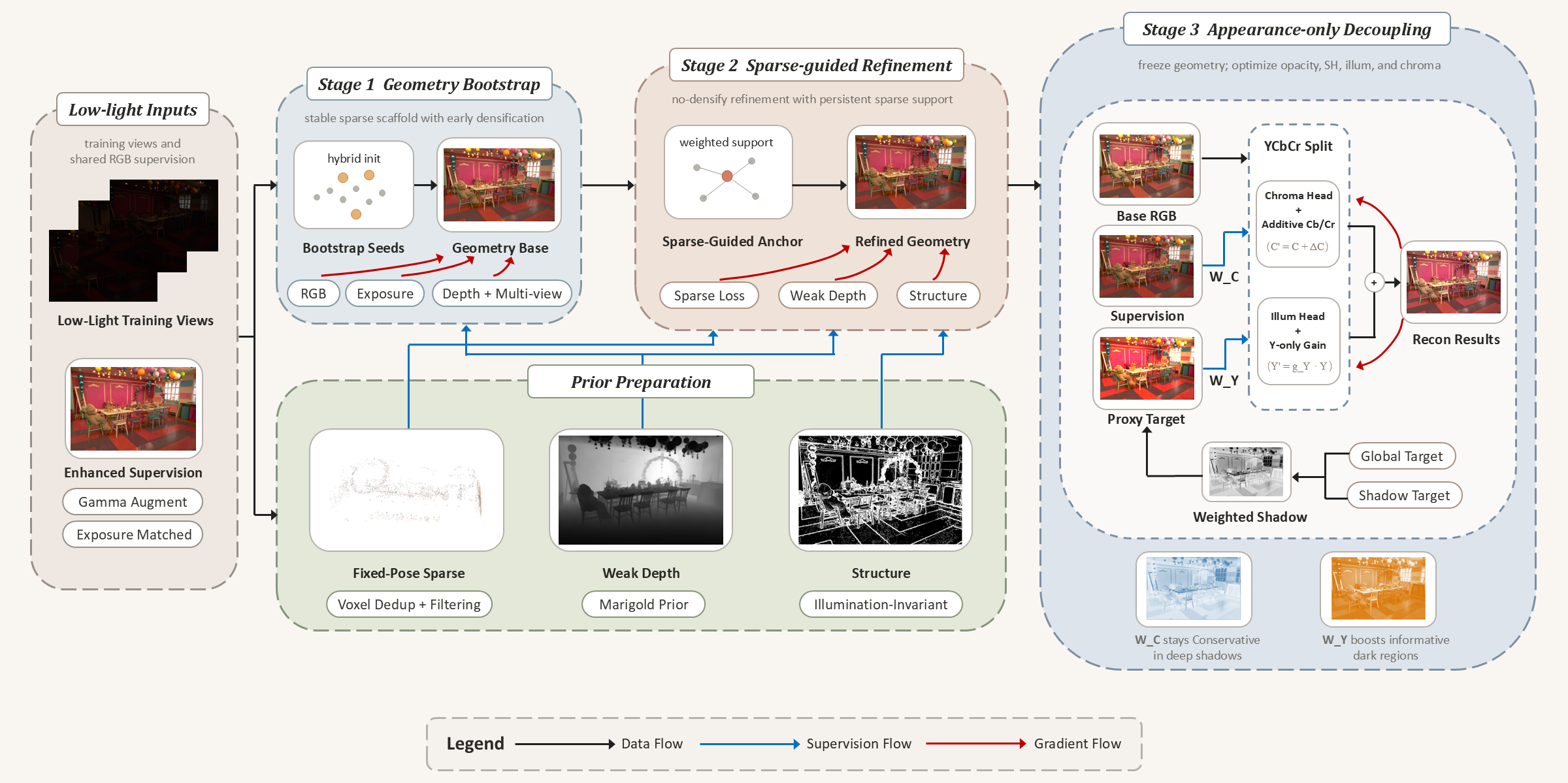}
\caption{Overview of SLL-GS~\cite{guo2026reliability}. A three-stage low-light 3DGS pipeline: Stages 1–2 build geometry from COLMAP anchors with depth and reprojection priors; Stage 3 restores appearance via an illumination head and a YCbCr chroma residual.}
\label{fig:sll-gs}
\end{figure}


\noindent\textbf{Method}~~
The authors propose a three-stage low-light 3D Gaussian Splatting (3DGS) pipeline~\cite{guo2026reliability} that decouples luminance recovery from chroma correction and relies on multi-view sparse consistency. Stage 1 initializes Gaussians using triangulated sparse points from a fixed-pose COLMAP model alongside random points. Training is supervised by RGB reconstruction with SSIM, weak exposure regularization, a Marigold depth prior, and weak multiview reprojection consistency, with densification restricted to the early period. Stage 2 refines the geometry without densification, reusing COLMAP sparse points as weak anchors via a Charbonnier-style robust loss and a \(k\)-NN neighborhood for smooth regularization. Finally, Stage 3 restores appearance using an adaptively calibrated proxy relighting target, confidence-aware reconstruction weighting, a scalar illumination head for brightness recovery, and an additive YCbCr chroma residual for color correction.\\


\noindent\textbf{Training Details}~~
The model is trained for 15,000 iterations in Stage 1, and 5,000 iterations each in Stages 2 and 3.
The initial Gaussian count is set to 100,000, with a spherical harmonics (SH) degree of 3 and a scene scale of 2.0. Pretrained Marigold is utilized solely for monocular depth priors, and no external paired relighting datasets are used. 

\subsection{GammaGS: Gamma-Calibrated 3D Gaussian Splatting with Affine Color Correction}
\label{subsec:HangFans}

\begin{center}


\noindent\emph{Wei Zhou$^{1}$, Linfeng Li$^{1}$, Qi Xu$^{2}$, Hang Song$^{3}$}


\noindent\emph{$^{1}$National University of Singapore, $^{2}$Shanghai Jiao Tong University, $^{3}$Xi'an Jiaotong University}

\end{center}

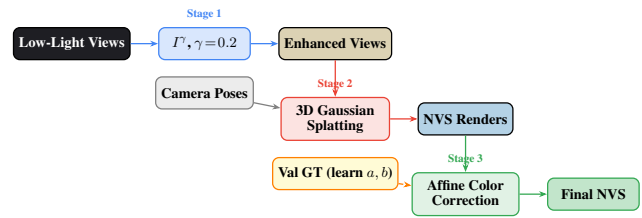
\begin{figure}[h]
    \centering
    \resizebox{\linewidth}{!}{%
    \begin{tikzpicture}[
        node distance=0.4cm,
        block/.style={rectangle, rounded corners=4pt, minimum width=2.0cm, minimum height=0.7cm, text centered, font=\small\bfseries, draw, thick},
        arrow/.style={-{Stealth[length=5pt]}, thick},
        label/.style={font=\scriptsize, midway, fill=white, inner sep=1pt},
    ]
    \node[block, fill=darkbg, text=white] (input) {Low-Light Views};
    \node[block, fill=stage1!20, draw=stage1, right=0.6cm of input] (gamma) {\(I^{\gamma}\), \(\gamma\!=\!0.2\)};
    \node[block, fill=brightbg, right=0.6cm of gamma] (enhanced) {Enhanced Views};

    \node[block, fill=stage2!15, draw=stage2, below=0.8cm of enhanced] (gs) {\begin{tabular}{c}3D Gaussian\\[-2pt] Splatting\end{tabular}};
    \node[block, fill=nvsbg, right=0.6cm of gs] (nvs) {NVS Renders};

    \node[block, fill=gray!15, draw=gray, below=0.35cm of gamma] (poses) {Camera Poses};

    \node[block, fill=stage3!15, draw=stage3, below=0.8cm of nvs] (affine) {\begin{tabular}{c}Affine Color\\[-2pt] Correction\end{tabular}};
    \node[block, fill=stage3!30, draw=stage3, thick, right=0.6cm of affine] (final) {\textbf{Final NVS}};

    \node[block, fill=yellow!20, draw=orange, below=0.35cm of gs] (valgt) {Val GT (learn \(a,b\))};

    \draw[arrow, stage1] (input) -- (gamma);
    \draw[arrow, stage1] (gamma) -- (enhanced);
    \draw[arrow, stage2] (enhanced) -- (gs);
    \draw[arrow, gray] (poses) -- (gs);
    \draw[arrow, stage2] (gs) -- (nvs);
    \draw[arrow, stage3] (nvs) -- (affine);
    \draw[arrow, stage3] (affine) -- (final);
    \draw[arrow, orange, dashed] (valgt) -- (affine);

    \node[font=\scriptsize\bfseries, stage1, above=0.05cm of gamma] {Stage 1};
    \node[font=\scriptsize\bfseries, stage2, above=0.05cm of gs] {Stage 2};
    \node[font=\scriptsize\bfseries, stage3, above=0.05cm of affine] {Stage 3};

    \end{tikzpicture}%
    }
    \caption{\textbf{GammaGS pipeline.} Stage 1: gamma correction restores brightness. Stage 2: 3DGS reconstruction. Stage 3: per-channel affine transform corrects systematic color biases.}
    \label{weichow:pipeline}
\end{figure}


\noindent\textbf{Method}~~
The authors propose GammaGS, a three-stage pipeline designed to address extreme under-exposure and systematic color biases. 
(1) \textbf{Brightness Restoration}: A uniform gamma correction (\(I_{\text{enh}} = I_{\text{low}}^{\gamma}\) with \(\gamma=0.2\)) is applied to all training images. 
(2) \textbf{3DGS Reconstruction}: A 3DGS model is trained directly on the gamma-corrected views. 
(3) \textbf{Affine Color Calibration}: To correct systematic per-channel color biases caused by the mismatch between enhanced training views and normal-light GT, a per-channel affine transform (\(I_{\text{GT}}^{(c)} = a_c \cdot I_{\text{render}}^{(c)} + b_c\)) is learned via least-squares and applied to the final renders. The overall pipeline is illustrated in Fig.~\ref{weichow:pipeline}.\\

\noindent\textbf{Training Details}~~
It initializes with 100K random Gaussians and is trained for 30K steps. The loss function combines L1 and SSIM losses (\(\lambda_{\text{SSIM}} = 0.2\)). Densification occurs between steps 500 and 15K with a gradient threshold of \(2 \times 10^{-4}\), and SH degree is set to 3. The affine color calibration parameters, learned on the validation scene, are fixed as follows: \(a_R = 1.74\), \(b_R = -49.8\); \(a_G = 1.40\), \(b_G = -39.0\); \(a_B = 0.59\), \(b_B = 27.7\).

\subsection{DarkIR-GS}
\label{subsec:fjnustar}

\begin{center}


\noindent\emph{Chenkun Guo$^{1}$, Yufei Li$^{1}$, Zhanqi Shi$^{1}$}

\noindent\emph{$^{1}$Fujian Normal University}
\end{center}

\vspace{-1em}

\begin{figure}[h]
\centering
\includegraphics[width=\linewidth]{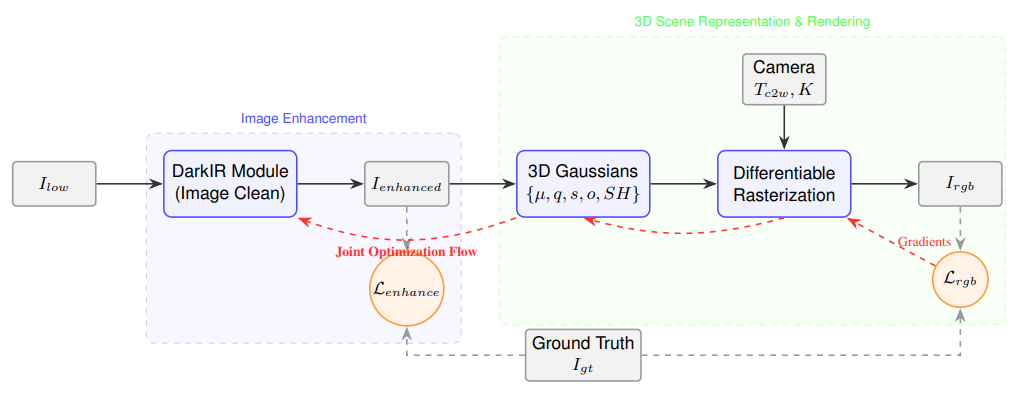}
\caption{Overview of DarkIR-GS. A DarkIR module restores illumination from low-light inputs, and differentiable rasterization back-propagates the RGB reconstruction loss into both the 3DGS parameters and the enhancement module for end-to-end training.}
\label{fjnu-star}
\end{figure}


\noindent\textbf{Method}~~
The authors propose DarkIR-GS, an end-to-end joint optimization framework integrating image enhancement and 3D reconstruction, as shown in Fig. \ref{fjnu-star}. A lightweight encoder-decoder network first restores illumination from low-light inputs. These enhanced images are then processed by 3DGS \cite{kerbl20233d} module. Differentiable rasterization allows gradients to flow back directly to the enhancement module. The total loss combines \( L_1 \) RGB reconstruction and pixel-wise enhancement losses.

\subsection{3DLLR}
\label{subsec:stamina}

\begin{center}


\noindent\emph{Bowen He$^{1,2}$, Qiang Zhu$^{2}$, Hantang Li$^{3}$, Xiandong Meng$^{2}$\\ Gang He$^{1}$}


\noindent\emph{$^{1}$Xidian University, $^{2}$Pengcheng Laboratory, $^{3}$Harbin Institute of Technology}

\end{center}

\begin{figure}[t]
\centering
\includegraphics[width= \linewidth]{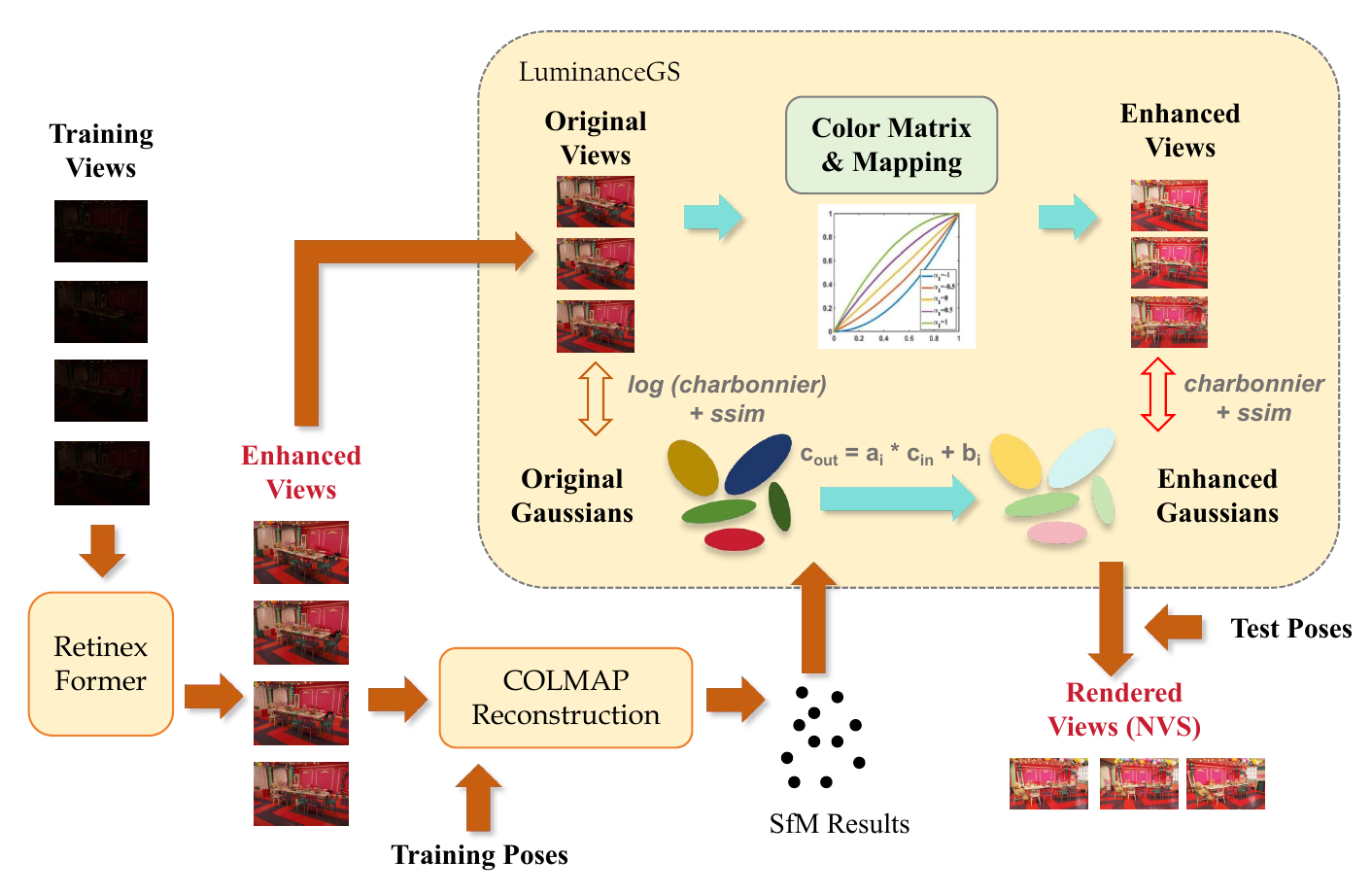}
\caption{The overall framework of 3DLLR.}
\label{3dllr_pipeline}
\end{figure}

\noindent\textbf{Method}~~
The authors propose a two-stage pipeline for low-light 3D reconstruction, as illustrated in Fig.~\ref{3dllr_pipeline}. First, input views are enhanced using RetinexFormer \cite{cai2023retinexformer}. These enhanced images are then processed via COLMAP with adjusted parameters and geometric consistency filtering. Then, the Luminance-GS \cite{cui2025luminance} training pipeline is adopted. The original optimization objective of Luminance-GS is modified by replacing the default loss with a combination of log Charbonnier and SSIM losses.\\

\noindent\textbf{Training Details}~~
The model is trained for 10,000 iterations with a batch size of 1 using the Adam optimizer (\(\beta_1 = 0.9\), \(\beta_2 = 0.999\)). An exponential scheduler decays learning rates to 1\% of their initial values. 
The total loss is formulated as:
\(
\mathcal{L}_{total} = \mathcal{L}_{low} + 0.5 \cdot \mathcal{L}_{enh} + \mathcal{L}_{spa} + 10 \cdot \mathcal{L}_{hist}
\)
where \(\mathcal{L}_{low}\) is the low-light reconstruction loss (log(Charbonnier) + SSIM, \(\lambda_{SSIM} = 0.2\)), \(\mathcal{L}_{enh}\) is the enhanced image reconstruction loss, \(\mathcal{L}_{spa}\) is the spatial consistency loss, and \(\mathcal{L}_{hist}\) is the histogram prior loss.
Gaussian densification occurs every 100 iterations (steps 500 to 8000), with opacities reset every 3000 iterations. The thresholds are set as follows: 0.005 for prune opacity, 0.0002 for splitting gradient, 0.01 for duplication scale, and 0.1 for pruning scale. The SH degree is 3, progressively increasing by one every 1000 iterations until reaching the full degree at step 3000.



\subsection{LLE-GS: A Unified Reconstruction Pipeline Combining Lowlight Restoration with 3DGS}
\label{subsec:AAA}

\begin{center}


\noindent\emph{Dufeng Zhang$^{1}$, Weizhi Nie$^{1}$, Weijie Wang$^{1}$, Xingan Zhan$^{1}$}


\noindent\emph{$^{1}$Tianjin University}

\end{center}

\vspace{-1em}

\begin{figure}[h]
    \centering
    \includegraphics[width=\linewidth]{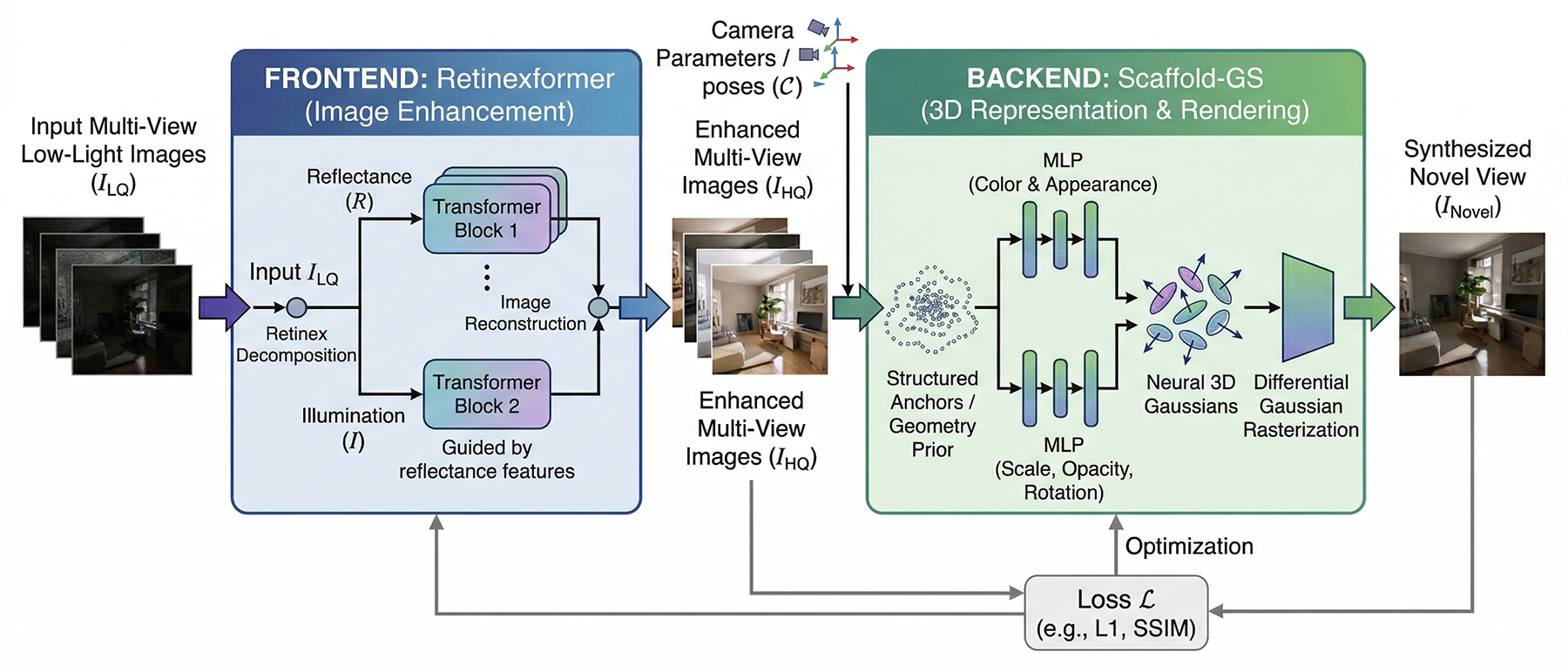}
    \caption{Overview of LLE-GS. An end-to-end framework that uses a transformer-based Retinex decomposition frontend to extract visual priors from low-light multi-view inputs, driving a view-adaptive Scaffolded 3DGS backend for high-quality NVS.}
    \label{LLE-GS:architecture}
\end{figure}


\noindent\textbf{Method}~~
The authors propose \textbf{LLE-GS}, a joint pipeline combining a low-light image restoration front-end with a view-adaptive 3D reconstruction back-end (Fig.~\ref{LLE-GS:architecture}). The front-end adopts a One-stage Retinex-based Framework utilizing an Illumination-Guided Transformer and a self-attention mechanism. This module estimates illumination and repairs degradations to output high-quality multi-view images. The back-end constructs a hierarchical, region-aware 3D representation using SfM points. Visible anchors dynamically decode neural Gaussians via MLPs, conditioned on local features, viewing distance, and relative direction, enabling robust view-adaptive rendering.\\


\noindent\textbf{Training Details}~~
Front-end training data is augmented with random rotation, horizontal flipping, and $128 \times 128$ crops. It is optimized using Adam with a cosine annealing schedule to minimize Mean Absolute Error (MAE). 

The resulting point cloud is voxelized to extract voxel centers as initialized anchors. Back-end anchors and MLPs are optimized end-to-end using $\mathcal{L}_1$, SSIM, and volume regularization losses. Additionally, an error-driven strategy dynamically grows anchors based on accumulated gradients and prunes those with low opacity.


\subsection{RunAI-Harmony3D}
\label{subsec:runai}

\begin{center}
\noindent\emph{Runyi Yang$^{1}$, Deheng Zhang$^{1}$, Yuqian Fu$^{2}$} \\

\noindent\emph{$^{1}$INSAIT, $^{2}$KAUST}
\end{center}
\vspace{-1.5em}
\begin{figure}[h]
    \centering
    \includegraphics[width=\linewidth]{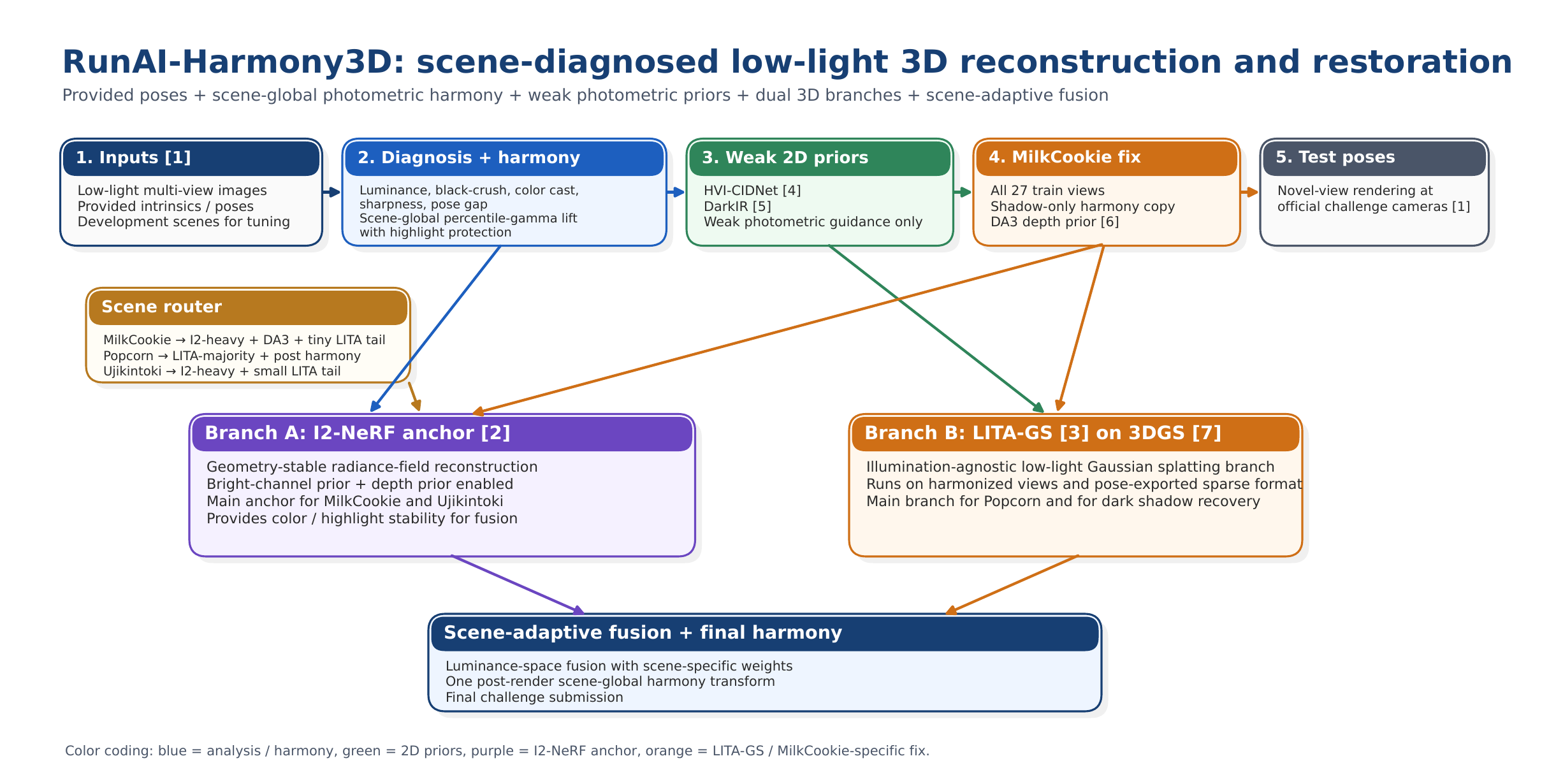}
    \caption{Overview of RunAI-Harmony3D.}
    \label{fig:runai_overview}
\end{figure}

\noindent\textbf{Method}~~
The authors propose RunAI-Harmony3D, which utilizes global photometric references and two complementary 3D branches: I2-NeRF \cite{liu2025i2nerf} (geometry-stable anchor) and LITA-GS \cite{zhou2025lita} (low-light recovery). HVI-CIDNet \cite{hvi}, DarkIR \cite{darkir}, and DA3 \cite{depthanything3} are integrated to provide 2D photometric and pose-conditioned depth priors. The pipeline consists of 4 stages: (1) scene diagnostics,  (2) scene-global photometric harmonization, (3) weak priors creation, and (4) dual-branch optimization. Final renderings are generated via luminance-space scene-adaptive fusion followed by a post-render harmony transform.\\

\noindent\textbf{Training Details}~~
Default optimization settings for I2-NeRF and LITA-GS are maintained. Development scenes are utilized to tune routing weights and harmony strength, and all official training views are used for blind-scene runs. For the challenging MilkCookie scene, original low-light RGB images are kept for optimization, while a shadow-only harmonized copy is generated specifically for the DA3 depth prior, restricting LITA-GS to the darkest flat regions.

\subsection{AIC-GER: Adaptive Illumination Correction and Global Exposure Regularization for Gaussian Splatting}
\label{subsec:cc}

\begin{center}


\noindent\emph{Cunchuan Huang$^{1}$ \quad $^{1}$Information Engineering University}

\end{center}

\vspace{-1.5em}
\begin{figure}[h]
    \centering
    \includegraphics[width=\linewidth]{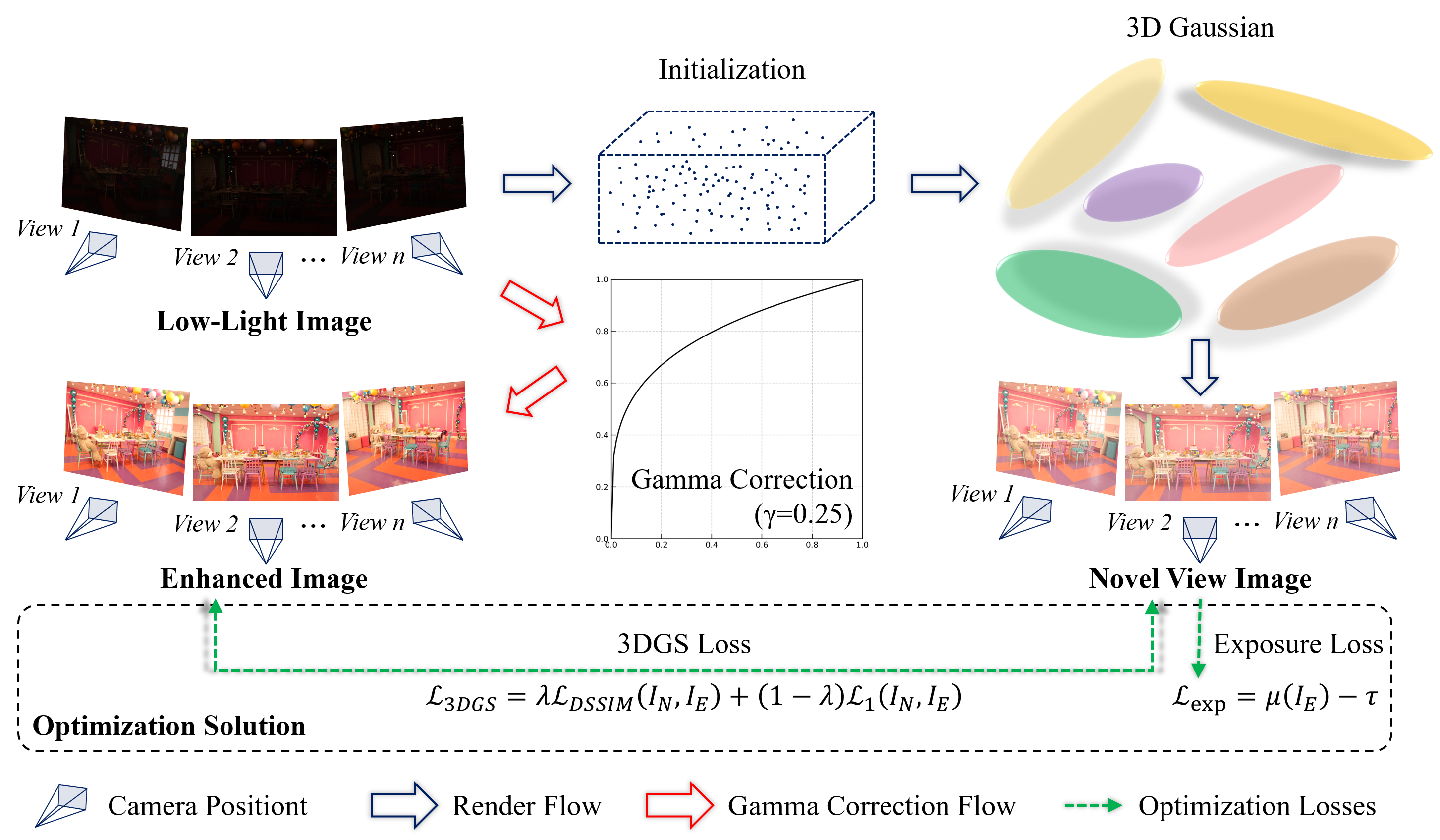}
    \caption{Overall pipeline of the proposed AIC-GER method.}
    \label{fig:cc_pipeline}
\end{figure}


\noindent\textbf{Method}~~
The authors propose AIC-GER, which builds upon the provided baseline to improve robustness under severe low-light degradations (see Fig.~\ref{fig:cc_pipeline}). It applies an enhanced low-light correction strategy via gamma adjustment to preprocess multi-view degraded inputs. To mitigate inconsistent exposure across views and regulate global brightness, an unsupervised global exposure regularization loss inspired by Aleth-NeRF \cite{cui2024aleth} is used. Specifically, the global mean luminance of the rendered image is enforced to approximate a predefined target value: 
\(
\mathcal{L}_{exp} = \left( \mu(I) - \tau \right)^2
\)
where \(\mu(I)\) denotes the global average luminance.\\ 


\noindent\textbf{Training Details}~~
For the improved low-light enhancement, a more aggressive gamma correction with a coefficient of 0.25 is adopted (compared to the baseline's 0.5). 
During training, the 3D reconstruction objective is jointly optimized with the proposed global exposure regularization, where the target exposure level \(\tau\) is set to 0.5.


\section{Track 2: 3D Smoke Restoration}
\label{subsec:t2_methods}

\subsection{GenSmoke-GS}
\label{subsec:plbbl}

\begin{center}


\noindent\emph{Qida Cao$^{1}$, Xinyuan Hu$^{1}$, Changyue Shi$^{1}$, Jiajun Ding$^{1}$\\ Zhou Yu$^{1}$, Jun Yu$^{2}$}


\noindent\emph{$^{1}$Hangzhou Dianzi University, $^{2}$Harbin Institute of Technology (Shenzhen)}

\end{center}

\begin{figure}[tbp]
    \centering
    \includegraphics[width=\linewidth]{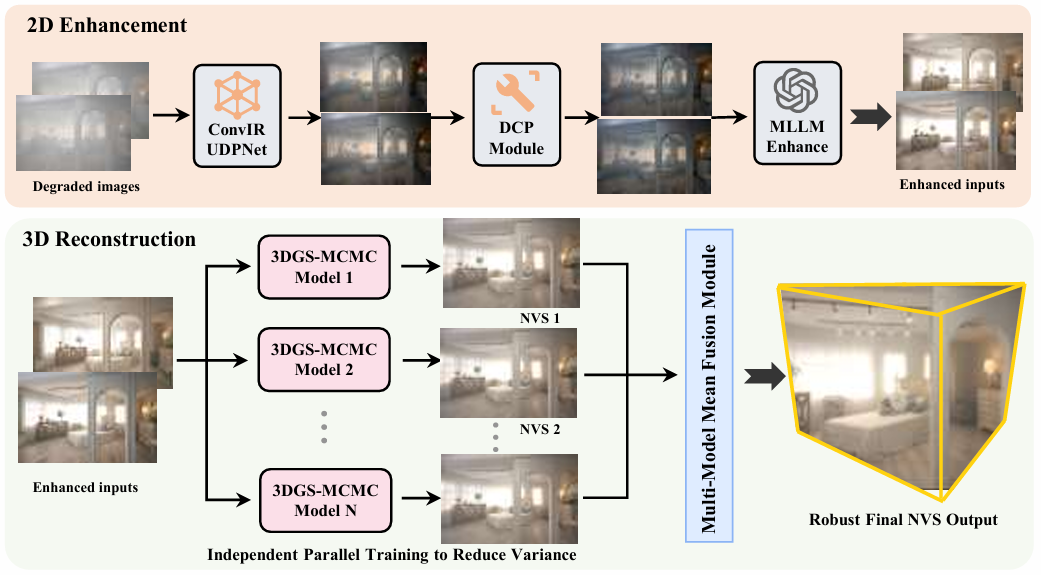}
    \caption{Overview of GenSmoke-GS~\cite{cao2026gensmoke}. Smoke-degraded multi-view images are sequentially restored via ConvIR-UDPNet, DCP-based dehazing, illumination correction, and MLLM-guided refinement under structural constraints, then reconstructed using 3DGS-MCMC with a FasterGS backend with ensembling.}
    \label{fig:pipeline_plbbl}
\end{figure}

\noindent\textbf{Method}~~
The authors propose GenSmoke-GS~\cite{cao2026gensmoke}, a multi-stage framework integrating restoration, physically grounded enhancement, and controlled MLLM-based generation (see Fig.~\ref{fig:pipeline_plbbl}). Given smoke-degraded multi-view images, ConvIR-UDPNet~\cite{cov,zuo2026udpnet} is initially applied for preliminary restoration to recover low-level structures and color information. Next, Dark Channel Prior~\cite{dcp} is utilized to remove residual haze and adjust global illumination. Subsequently, reconstruction-oriented MLLM enhancement is performed on each view independently. The prompt explicitly instructs the MLLM to preserve original geometry, layout, and scene structure, focusing solely on improving visibility and refining local details to mitigate structural drift. 
The enhanced images are then processed by a 3DGS-MCMC~\cite{mcmc} pipeline with a FasterGS~\cite{hahlbohm2026faster} backend. The maximum number of Gaussians is constrained to encourage cross-view consistency and suppress hallucinated structures. Finally, to reduce the variance inherent in optimization-based reconstruction, the pipeline is executed independently $n$ times, and the rendered NVS outputs are averaged to yield robust final predictions.\\



\noindent\textbf{Training Details}~~
GPT-Image-1.5~\cite{openai2025gptimage} is employed as the generative model during the MLLM enhancement stage. For the reconstruction stage, the 3DGS-MCMC~\cite{mcmc} pipeline is configured with a maximum of 100k Gaussians and optimized for 30k iterations. $n=91$ for ensembling. 

\subsection{Smoke-GS}
\label{subsec:xinsight_lab_t2}

\begin{center}


\noindent\emph{Fei Wang$^{1,2}$, Zhiliang Wu$^{2}$, Xinye Zheng$^{1}$, Kun Li$^{4}$\\ Yanyan Wei$^{1}$, Weisi Lin$^{3}$}


\noindent\emph{$^{1}$Hefei University of Technology, $^{2}$Hefei Comprehensive National Science Center, $^{3}$Nanyang Technological University, $^{4}$United Arab Emirates University}

\end{center}

\begin{figure}[tbp]
\centering
\includegraphics[width=\linewidth]{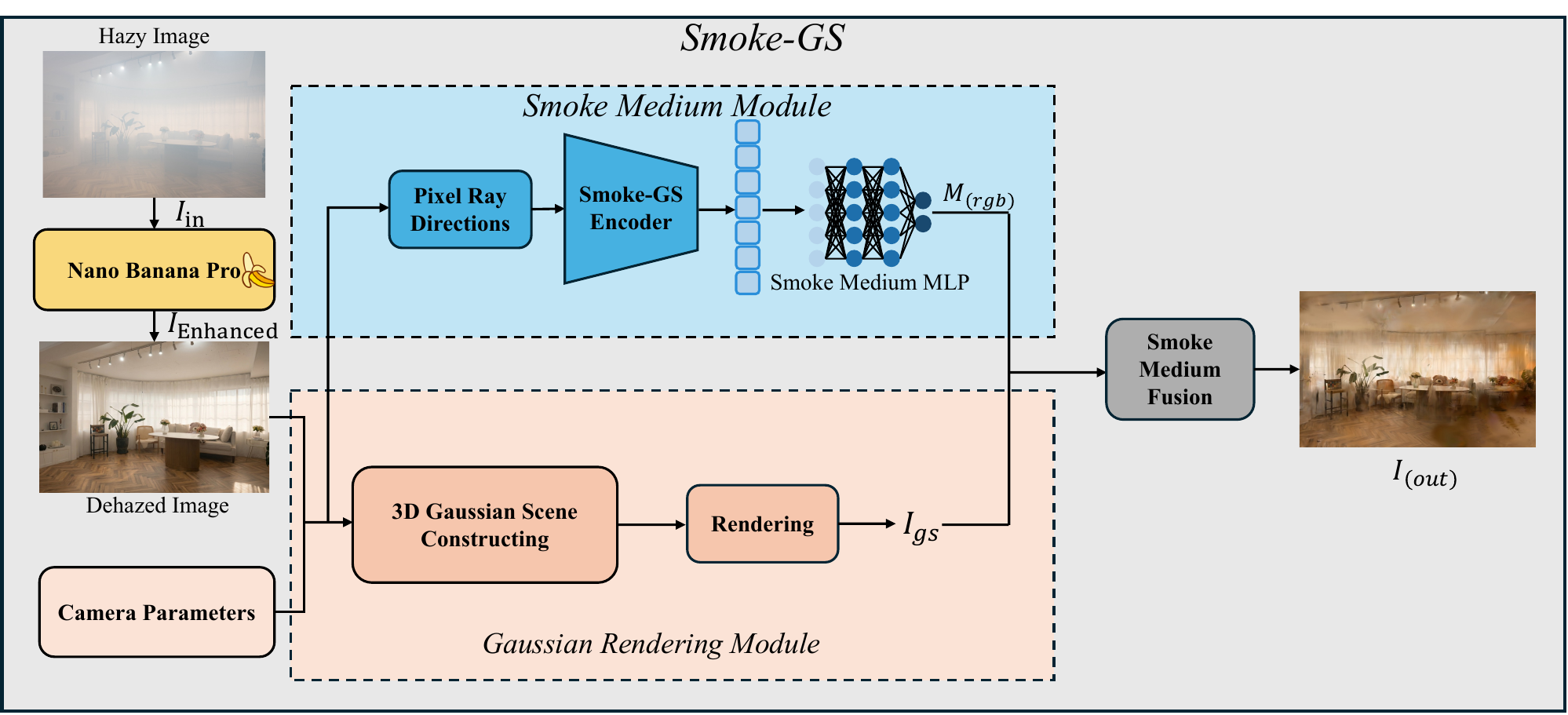}
\caption{Overview of Smoke-GS~\cite{zheng20263d}. Hazy images are first enhanced by NanoBanana Pro, then a Smoke Medium Module encodes pixel ray directions via the Smoke-GS Encoder and predicts medium parameters through a Smoke Medium MLP. Finally, the predicted medium terms are fused with 3DGS rendering.}
\label{fig:xinsight_t2}
\end{figure}

\noindent\textbf{Method}~~
The authors propose Smoke-GS~\cite{zheng20263d}, as illustrated in Fig.~\ref{fig:xinsight_t2}. First, Nano Banana Pro is utilized for image restoration to preserve geometric and texture details while enhancing contrast and brightness. To model the view-dependent appearance changes caused by scattering media such as smoke, a Smoke Medium Module is introduced. Based on the camera intrinsics and poses of the target view, the corresponding viewing ray direction for each pixel is constructed. These direction vectors are encoded using Spherical Harmonics in the Smoke-GS Encoder, followed by a Smoke Medium MLP that outputs 9-dimensional medium parameters divided into \textit{medium\_rgb}, \textit{medium\_bs}, and \textit{medium\_attn}. The \textit{medium\_rgb} component captures the view-dependent color shift introduced by the medium. Sigmoid is applied to \textit{medium\_rgb}, and this color correction term is combined with the base 3D Gaussian-rendered result to generate the final predicted image.\\

\noindent\textbf{Training Details}~~
For each scene, 100,000 Gaussians are initialized alongside the Smoke Medium MLP. The models are trained for 150,000 steps using the Adam optimizer. The initial learning rates for position, rotation, scale, and opacity are set to 1.6e-4, 1e-3, 5e-3, and 5e-2, respectively, while the learning rate for the MLP is 1e-3. The loss function is a weighted combination of L1 and SSIM, defined as \( L = (1 - \lambda)L1 + \lambda(1 - \text{SSIM}) \), with \( \lambda = 0.2 \).


\subsection{Dehaze-then-Splat: Generative Dehazing with Physics-Informed GS for Novelview Synthesis}
\label{subsec:hunanduo}

\begin{center}


\noindent\emph{Yuchao Chen$^{1}$, Hanqing Wang$^{2}$}


\noindent\emph{$^{1}$Huazhong University of Science and Technology, $^{2}$Hong Kong University of Science and Technology (Guangzhou)}

\end{center}


\noindent\textbf{Method}~~
The authors present a two-stage dehaze-then-reconstruct pipeline~\cite{chen2026dehaze}. In the first stage, pseudo-clean training images are produced via frame-wise generative dehazing using Nano Banana Pro. To enforce multi-view consistency and correct frame-wise brightness shifts, per-channel mean/std brightness normalization is applied to the dehazed outputs, which are subsequently resized to match the camera intrinsics. 
In the second stage, the scene is reconstructed using 3DGS augmented with physics-informed priors. The model is optimized using a composite loss function:
\(
\mathcal{L} = (1{-}\lambda_s)\mathcal{L}_1 + \lambda_s \mathcal{L}_\text{SSIM} + \lambda_d \mathcal{L}_\text{DCP} + \lambda_p \mathcal{L}_\text{depth} + \lambda_g \mathcal{L}_\text{grad}
\)
where $\mathcal{L}_\text{depth}$ is a scale-invariant weighted Pearson correlation loss utilizing pseudo-depth maps generated by Depth-AnythingV2 \cite{depth_anything_v2} to geometrically anchor Gaussian positions. $\mathcal{L}_\text{DCP}$ applies DCP regularization to the rendered images to suppress residual haze. $\mathcal{L}_\text{grad}$ provides structural guidance via Sobel gradient matching against MB-TaylorFormer \cite{qiu2023mb} dehazing output.\\ 


\noindent\textbf{Training Details}~~
Training is conducted for 20k steps with the SH degree progressively increased from 0 to 3, and a scene scale of 2.0. The learning rates are set to $\text{lr}_\text{means}{=}1.6{\times}10^{-4}$ and $\text{lr}_\text{SH0}{=}2.5{\times}10^{-3}$. The image augmentation parameter \texttt{GAMMA} is strictly set to 1.0 to avoid unintended nonlinear transforms.
For MCMC densification, \texttt{CAP\_MAX} is set to 500k and \texttt{NOISE\_LR} to $5{\times}10^5$ (decayed to 0 after step 8000), with densification running from step 500 to an early stop at step 3000. Most scenes are initialized with 50k random points and a white background, except for the dark indoor scene (Hinoki), which requires a reduced initialization of 20k points and a black background. Final predictions are generated from early checkpoints (typically steps 2000--4000) to maximize PSNR.


\subsection{MSDG: Multi-Stage Dehazing Gaussian}
\label{subsec:team_aiia_lab}

\begin{center}


\noindent\emph{Yang Gu$^{1}$, Jiacheng Liu$^{2}$, Shiyu Liu$^{1}$, Kui Jiang$^{1}$\\ Junjun Jiang$^{1}$}


\noindent\emph{$^{1}$Harbin Institute of Technology, $^{2}$Northeastern University}

\end{center}

\begin{figure}[!hp]
    \centering
    \includegraphics[width=\linewidth]{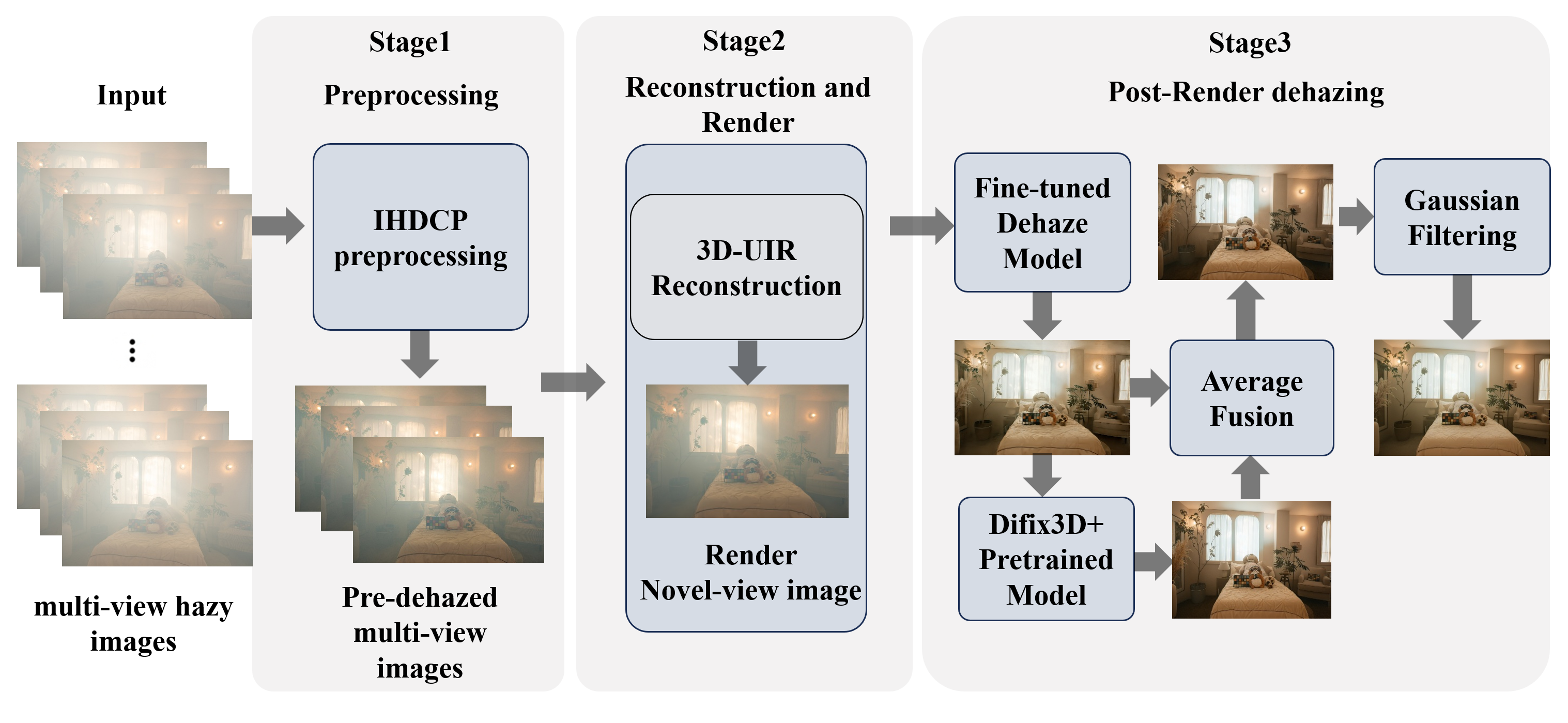}
    \caption{Overview of the MSDG pipeline. Multi-view hazy images are mildly pre-dehazed, fed into 3D-UIR for rendering, processed by a fine-tuned dehazing model, refined by Difix3D+, and finally fused and smoothed.}
    \label{fig:pipeline_aiia}
\end{figure}


\noindent\textbf{Method}~~
The authors propose a multi-stage pipeline for dehazed novel view reconstruction shown in Fig.~\ref{fig:pipeline_aiia}. First, an IHDCP \cite{liu2026ihdcp}-based mild pre-dehazing is applied to multi-view hazy images to suppress haze while preserving structural consistency. The preprocessed images are then fed into the 3D-UIR \cite{yuan2025_3duir} framework for robust 3D reconstruction and novel-view rendering. To further enhance visibility and correct color deviations, the rendered views are processed by a scene-specific fine-tuned dehazing model. A pretrained Difix3D+ \cite{wu2025difix3d+} model is subsequently utilized to repair local degraded regions and restore texture details. The final output is generated via mean fusion of the dehazed and repaired views, followed by Gaussian smoothing to suppress residual noise and fusion artifacts.\\ 


\noindent\textbf{Training Details}~~
During training, the 3D-UIR backbone is optimized using a composite objective comprising a pixel- and structural-level reconstruction loss (\(L_1\) and D-SSIM), a depth regularization loss to refine depth estimation using pseudo-depth, and a scale regularization loss that encourages 3DGS to concentrate around physical surfaces.

\subsection{DiT-IBGS}
\label{subsec:team3ddd}

\begin{center}

\noindent\emph{Dizhe Zhang$^{1}$, Meixi Song$^{1}$, Haoran Feng$^{1}$}

\noindent\emph{$^{1}$Insta360 research}

\end{center}

\begin{figure}[!hp]
\centering
\includegraphics[width=\linewidth]{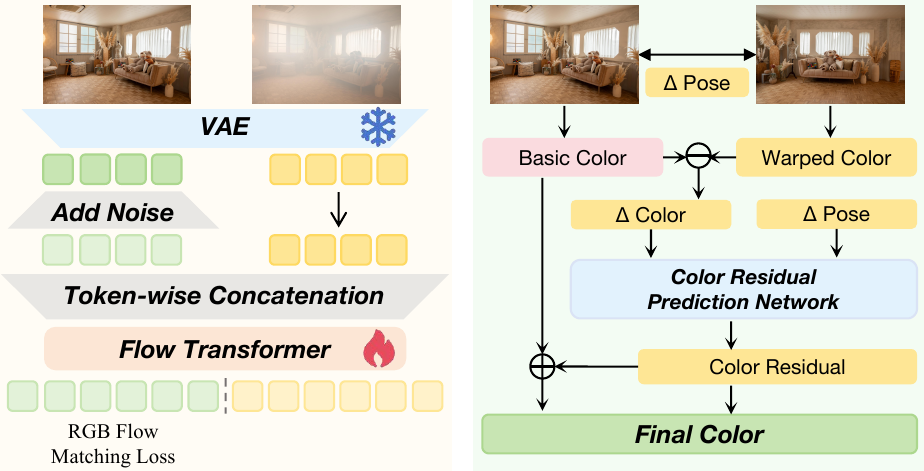}
\caption{Overview of the proposed DiT-IBGS.}
\label{network}
\end{figure}


\noindent\textbf{Method}~~
The authors propose a parameter-efficient latent diffusion framework based on the pretrained FLUX.1-dev model. As shown in Fig.~\ref{network}, degraded and target images are first VAE-encoded. Following the flow-matching formulation, the degraded latent is concatenated with the noisy target latent to serve as an explicit visual condition. A shifted positional offset is applied to distinguish these conditional tokens from the restoration branch. The transformer then processes the packed tokens alongside text embeddings to predict the clean target's latent flow. 
The enhanced views are subsequently fed into IBGS, which incorporates image-guided geometric and photometric constraints. To achieve geometry-aware rendering, the representation specifically learns normal and offset terms.\\ 



\noindent\textbf{Training Details}~~
The framework is implemented using \texttt{FLUX.1-dev}. The VAE and text encoders are frozen, while the transformer is fine-tuned using LoRA adapters (rank and alpha 64, dropout 0.1) on the attention and projection layers. Images are resized to \(592 \times 896\). Optimization is performed using AdamW (lr=\(5 \times 10^{-5}\)) for 50 epochs on 6 GPUs, utilizing a batch size of 3, gradient accumulation of 6, BF16 mixed precision, and DeepSpeed Stage-2. The guidance scale is set to 1.0, and the conditional positional offset is 10. A weighted MSE loss is applied between the predicted and ground-truth flow targets in the latent space.


\subsection{DEPHY-GS: Dehaze-Guided Hybrid 3DGS with Residual and Physical Modeling}
\label{subsec:team_diouj_el}

\begin{center}

\noindent\emph{Phan The Son$^{1}$ \quad $^{1}$VNU University of Science}

\end{center}

\begin{figure}[!hp]
    \centering
    \includegraphics[width=\linewidth]{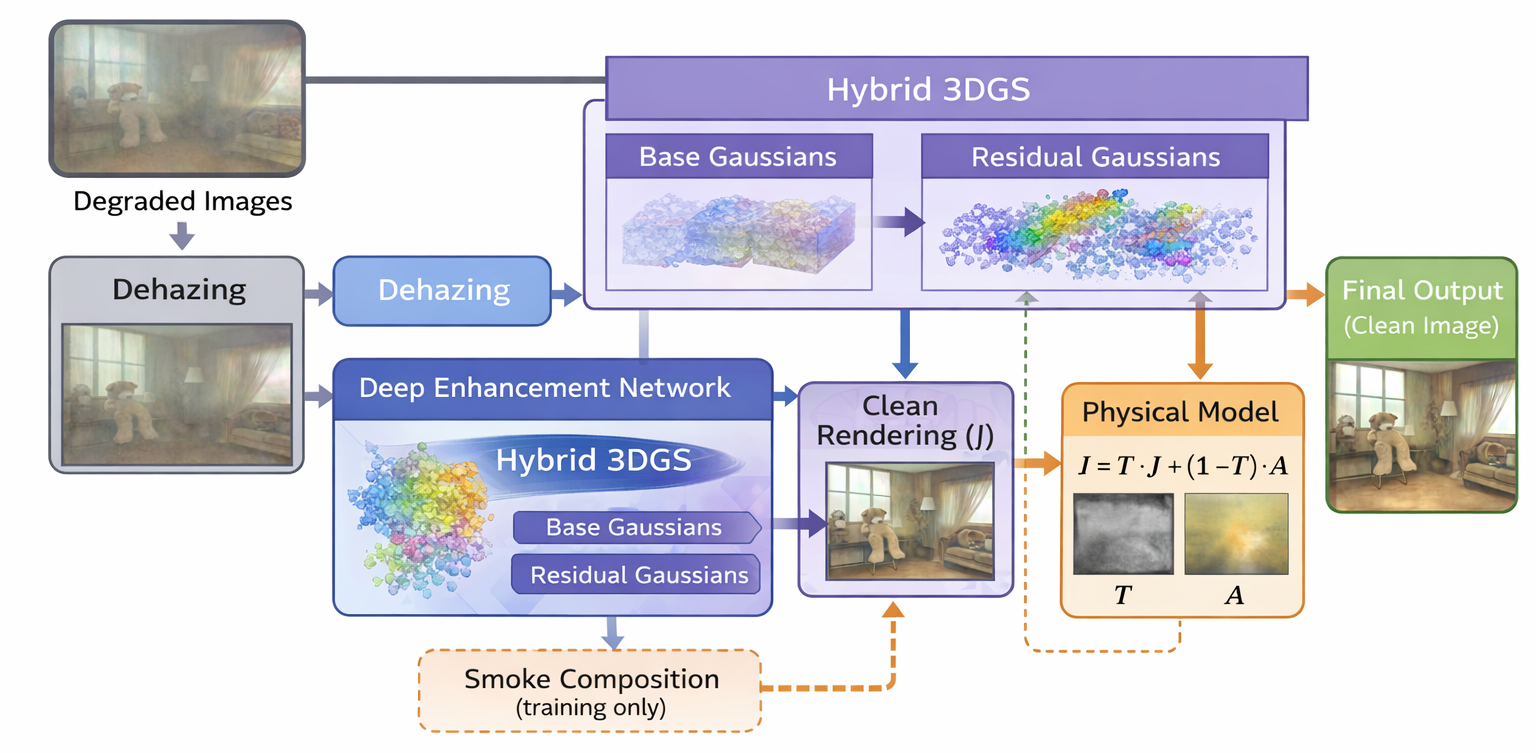}
    \caption{Overview of DePhy-GS. Smoke degradation is progressively removed via dehazing and deep enhancement, then reconstructed with a hybrid 3DGS representation and residual modeling, guided by a lightweight physical degradation model.}
    \label{fig:pipeline_diouj}
\end{figure}


\noindent\textbf{Method}~~
The authors propose a hybrid 3D-2D pipeline (see Fig. \ref{fig:pipeline_diouj}) consisting of three stages: 
First, a lightweight dehazing module conservatively suppresses the global veil effect. Second, a deterministic MLPs refines local details and reduces residual artifacts. Generative approaches (e.g., GANs or diffusion models) are intentionally avoided to maintain pixel-level fidelity. Third, the enhanced images optimize a 3DGS representation. A dual-layer representation is introduced: a base Gaussian set for primary scene structure and a residual Gaussian set for fine details and correction signals. The final clean rendering is obtained by combining both layers with a learnable residual gain.
To explicitly represent the degradation process, a lightweight 2D head is incorporated to predict transmission and airlight from the rendered clean image, following a physical image formation model:
\(
    I = T \cdot J + (1 - T) \cdot A,
\)
where \(I\) is the observed image, \(J\) is the clean scene radiance, \(T\) is the transmission map, and \(A\) is the airlight.

\subsection{SmokeGS-R: Physics-Guided 3DGS with Multi-Reference Color Harmonization}
\label{subsec:windrise}

\begin{center}


\noindent\emph{Xueming Fu$^{1}$, Lixia Han$^{2}$}


\noindent\emph{$^{1}$University of Science and Technology of China, $^{2}$Nanjing University of Aeronautics and Astronautics}

\end{center}

\begin{figure}[!htbp]
    \centering
    \includegraphics[width=\linewidth]{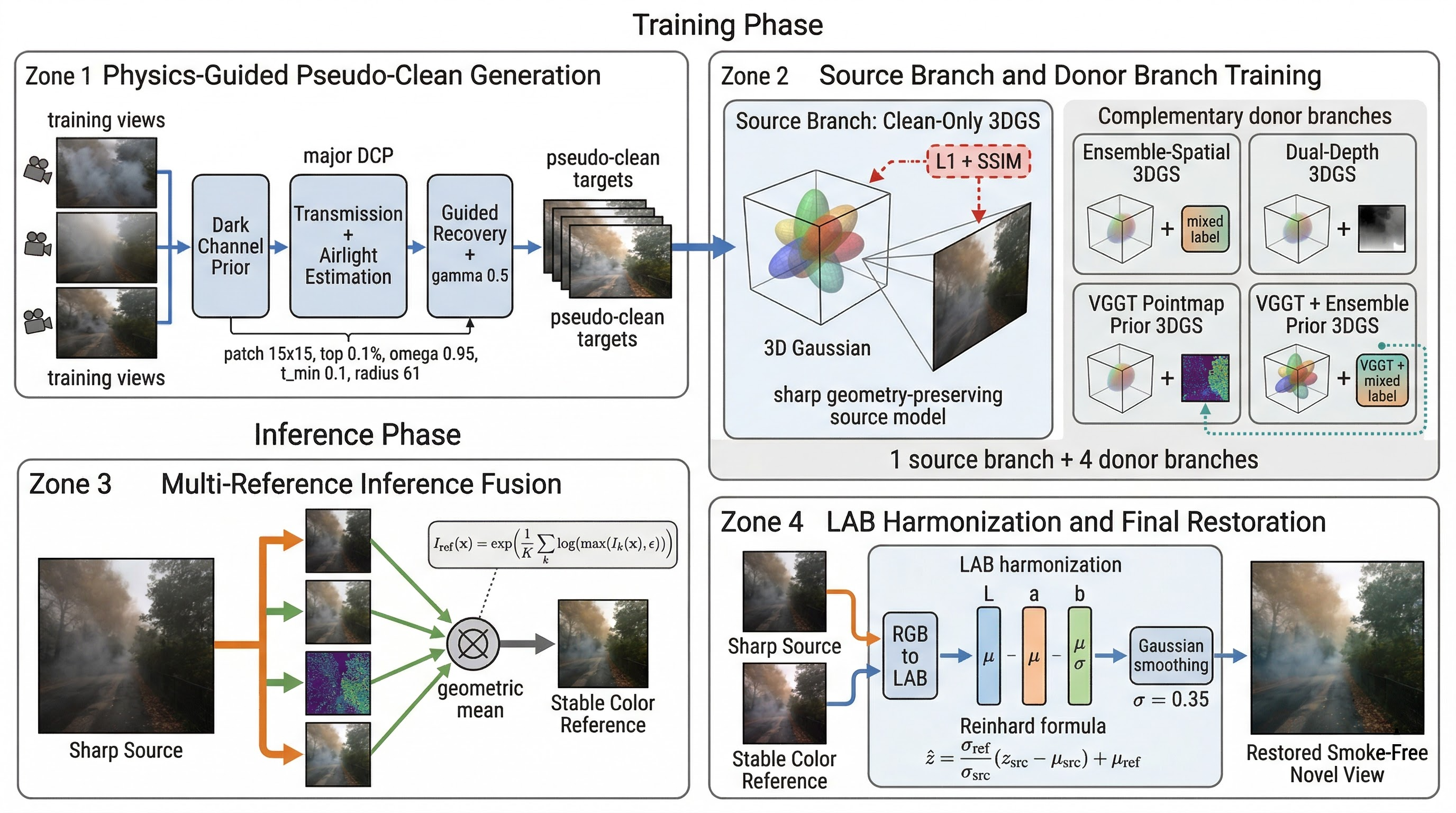}
    \caption{Overview of SmokeGS-R. Refined DCP inversion supervises a clean-only 3DGS with four donor branches as appearance priors; at inference, a geometric-mean reference from five renders harmonizes outputs in LAB space.}
    \label{fig:smokegs-r}
\end{figure}

\noindent\textbf{Method}~~
The authors propose SmokeGS-R~\cite{fu2026smokegs}, a three-stage pipeline decoupling structural reconstruction from color restoration (see Fig.~\ref{fig:smokegs-r}). First, physics-guided pseudo-clean targets are constructed from smoky inputs by inverting the atmospheric scattering model via a refined DCP \cite{dcp}. Second, a multi-prior 3DGS ensemble is trained, comprising a clean-only source branch for sharp geometry and donor branches (dual-depth, VGGT-prior, ensemble-spatial) for appearance robustness. Finally, perceptual color harmonization is performed via LAB-space ensemble transfer, combining source geometry with ensemble color statistics through Reinhard color transfer using a geometric-mean reference and Gaussian smoothing.\\


\noindent\textbf{Training Details}~~
Pseudo-clean generation utilizes dark channel computation, guided filtering (radius $61$, $\epsilon=10^{-3}$) for transmission estimation, and gamma correction ($\gamma=0.5$). The clean-only source 3DGS model is trained for 5,000 iterations with combined $\mathcal{L}_1$ and SSIM loss ($\lambda_{\text{ssim}}=0.2$). Donor models utilize complementary priors like monocular depth and VGGT pointmaps. 

\subsection{HAD-GS: Haze-Aware 3D Gaussian Reconstruction with Diffusion Refinement}
\label{subsec:zzz}

\begin{center}


\noindent\emph{Zhenyu Zhao$^{1}$, Zihan Zhai$^{1}$, Tingting Li$^{1}$}


\noindent\emph{$^{1}$Xidian University}

\end{center}


\noindent\textbf{Method}~~
They propose Haze-Aware 3D Gaussian Reconstruction with Diffusion Refinement. First, the authors preprocess the training dataset by removing part of the haze using a physical model based on the dark channel prior. Subsequently, they convert the multi-view images into 3D Gaussian parameters via the 3DRR method, perform differentiable rendering based on camera poses, and generate images from the desired viewpoints. Finally, they employ a diffusion-based dehazing model to progressively remove noise through diffusion sampling, recovering clear images.

\subsection{SDG-GS: Simplified Depth-Guided Smoke Forward Modeling with Clean 3DGS}
\label{subsec:team_dlmath_vision}

\begin{center}


\noindent\emph{Seungsang Oh$^{1}$, Donggun Kim$^{1}$, Jeongbin You$^{1}$, Younghyuk Kim$^{1}$, Il-Youp Kwak$^{2}$}


\noindent\emph{$^{1}$Korea University, $^{2}$Chung-Ang University}

\end{center}

\begin{figure}[tbp]
    \centering
    \includegraphics[width=0.9\linewidth]{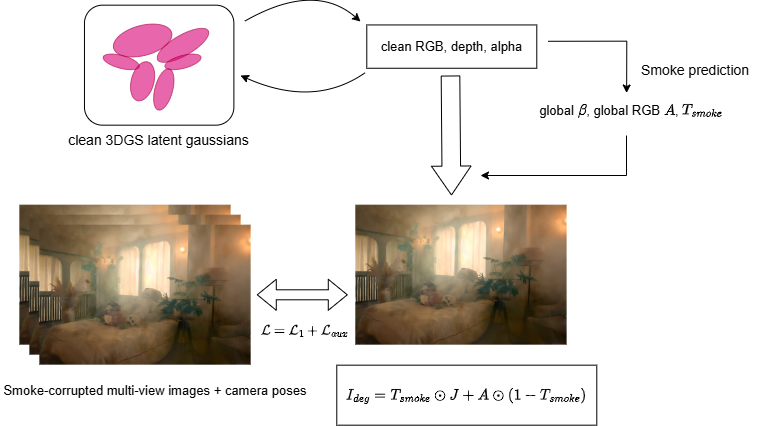}
    \caption{Overall architecture of SDG-GS. A clean 3D Gaussian scene representation is learned jointly with a simplified smoke forward model during training. At inference time, only the clean 3DGS renderer is used to produce the final restored test images.}
    \label{fig:architecture_dlmath_t2}
\end{figure}

\noindent\textbf{Method}~~
The restoration is formulated as a decomposition problem, separating the latent clean 3D scene from the smoke corruption process during joint optimization (Fig.~\ref{fig:architecture_dlmath_t2}). A 3DGS model renders a clean RGB image \(J\), depth map \(z\), and alpha map \(\alpha\). To prevent overfitting, a low-capacity convolutional predictor \(f_{\phi}\) estimates scene-level parameters—a global attenuation coefficient \(\beta\) and global RGB atmospheric light \(A\)—from \(J\), \(\alpha\), and normalized depth \(d\). 
Depth is preprocessed using a scene-consistent fixed normalization
with a fixed scale. 
Smoke transmission is modeled as \(T_{\text{raw}} = \exp(-\beta d)\) and clipped for stability: \(T_{\text{smoke}} = \mathrm{clip}(T_{\text{raw}}, 0.30, 1.0)\). The degraded image is synthesized via the atmospheric scattering model:
\( I_{\text{deg}} = T_{\text{smoke}} \odot J + A \odot (1 - T_{\text{smoke}}). \) \\




\noindent\textbf{Training Details}~~
The 3DGS and predictor parameters are jointly optimized using:
\(
\mathcal{L}_{\text{total}} = \lambda_1 \mathcal{L}_{\text{rec}} + \lambda_2 \mathcal{L}_{\text{grad}} + \lambda_3 \mathcal{L}_{\text{TV}}(T_{\text{smoke}}) + \lambda_4 \mathcal{L}_{\text{TV}}(\beta) + \lambda_5 \mathcal{L}_{\text{prior}}(A) + \lambda_6 \mathcal{L}_{\text{pseudo}}(T)
\)
Here, \(\mathcal{L}_{\text{rec}}\) is the pixel-wise reconstruction loss against smoke-corrupted inputs, \(\mathcal{L}_{\text{grad}}\) enforces gradient consistency, \(\mathcal{L}_{\text{TV}}\) terms ensure spatial smoothness for transmission and attenuation, \(\mathcal{L}_{\text{prior}}(A)\) regularizes atmospheric light, and \(\mathcal{L}_{\text{pseudo}}(T)\) applies a DCP to stabilize transmission. 

\subsection{EE-GS: Smoke-Robust 3D Reconstruction with Image Restoration and Depth-Guided Gaussian Splatting}
\label{subsec:eegs}

\begin{center}


\noindent\emph{Junbo Yang$^{1}$, Hongsen Zhang$^{1}$ \quad $^{1}$Xidian University}


\end{center}

\begin{figure}[!h]
    \centering
    \includegraphics[width=\linewidth]{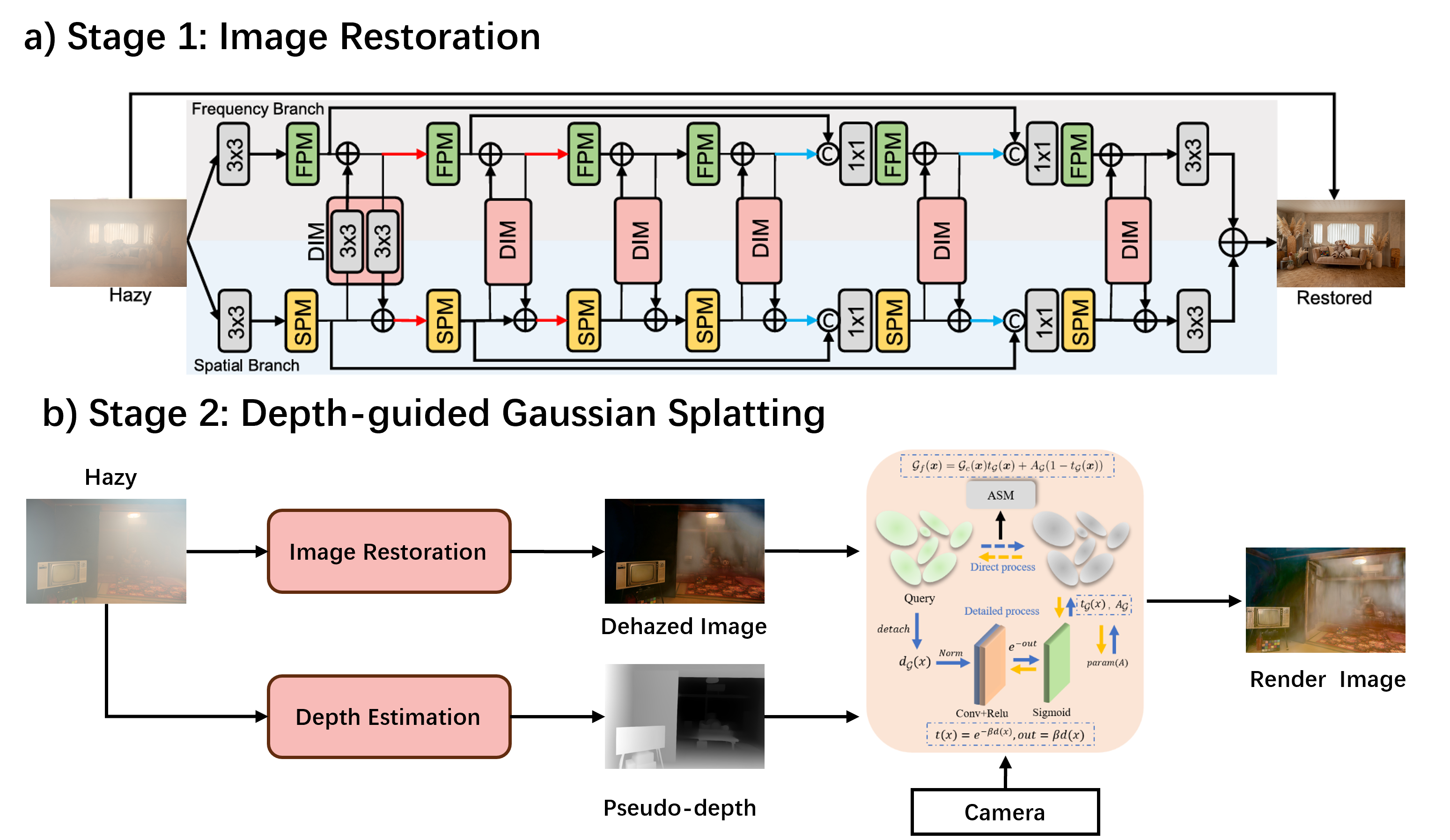}
    \caption{Overall architecture of EE-GS. Degraded views are first restored by EE-Net, and then processed by Depth-AnythingV2 to estimate dense depth maps. Finally, the restored images together with the depth estimates are used in 3DGS for scene optimization.}
    \label{fig:eegs}
\end{figure}

\noindent\textbf{Method}~~
The authors propose a restoration-guided pipeline combining EE-Net \cite{cui2025eenet}, Depth-AnythingV2 \cite{depth_anything_v2}, and 3DGS \cite{kerbl20233d}, as shown in Fig.~\ref{fig:eegs}. First, EE-Net processes smoke-degraded images to suppress scattering artifacts. Next, Depth-AnythingV2 estimates dense depth maps from these restored images to extract structural cues. Finally, the 3DGS representation is optimized using both the restored images and depth maps. The training is supervised by the restored observations, with the training objective defined as:
\(
\mathcal{L}_{\mathrm{rgb}} = (1-\lambda)\|\tilde{J}_i-\hat{J}_i\|_1 + \lambda \left(1-\mathrm{SSIM}(\tilde{J}_i,\hat{J}_i)\right),
\)
and $\lambda$ balances pixel-level fidelity and structural similarity.\\

\noindent\textbf{Training Details}~~
Training views are processed by EE-Net and Depth-AnythingV2 to generate the restored images and depth maps for standard 3DGS optimization.

\subsection{3DSmokeR}
\label{subsec:helicopter}

\begin{center}


\noindent\emph{Hantang Li$^{1,2}$, Qiang Zhu$^{2}$, Bowen He$^{3,2}$\\ Xiandong Meng$^{2}$, Debin Zhao$^{1}$, Xiaopeng Fan$^{1}$}


\noindent\emph{$^{1}$Harbin Institute of Technology (Shenzhen)\\$^{2}$Pengcheng Laboratory, $^{3}$Xidian University}

\end{center}

\vspace{-1em}
\begin{figure}[!h]
    \centering
    \includegraphics[width=\linewidth]{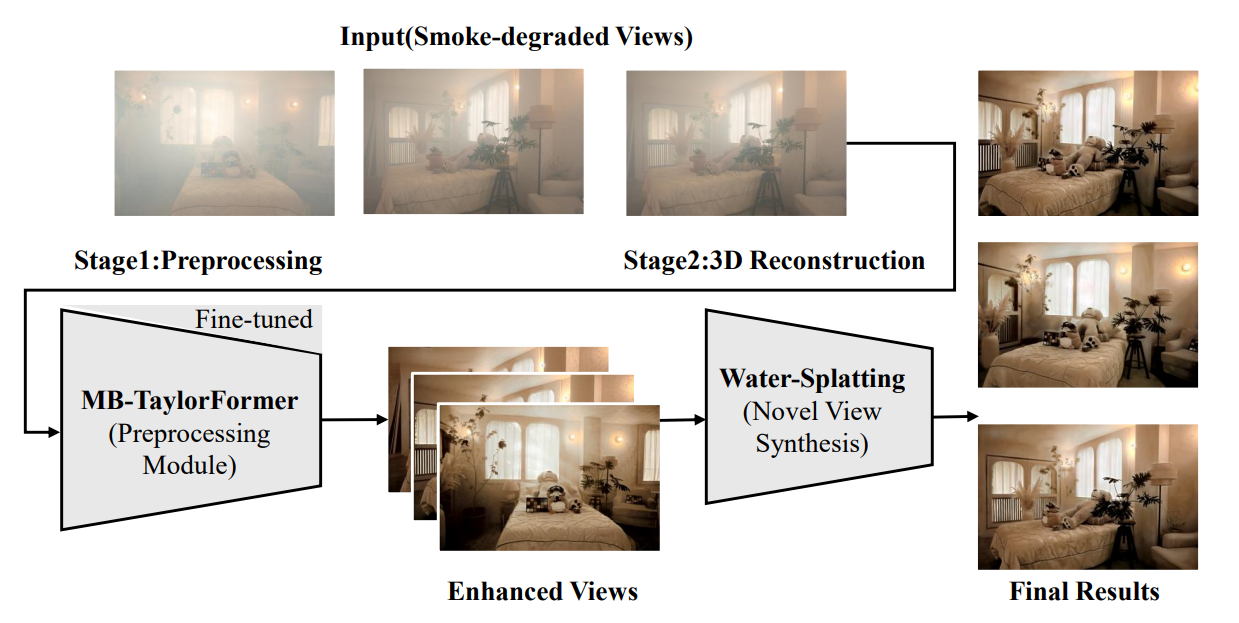} 
    \caption{Overview of 3DSmokeR. Stage 1 preprocesses smoke-degraded views with a fine-tuned MB-TaylorFormer; Stage 2 feeds the restored views into watersplatting \cite{li2025watersplatting} for 3D reconstruction.}
    \label{fig:helicopter}
\end{figure}

\noindent\textbf{Method}~~
The authors propose 3DSmokeR, a two-stage pipeline shown in Fig. \ref{fig:helicopter}. First, a fine-tuned MB-TaylorFormer \cite{qiu2023mb} acts as a preprocessing module to remove smoke-like degradation from input views. Then, the enhanced multi-view images are directly fed into Watersplatting \cite{li2025watersplatting} for 3D reconstruction following original 3DGS optimization.\\


\noindent\textbf{Training Details}~~
The MB-TaylorFormer \cite{qiu2023mb} is fine-tuned on a paired synthetic smoke dataset. Training employs a progressive strategy over four 5k-iteration phases, utilizing patch sizes \{128, 192, 256, 256\} and corresponding batch sizes \{4, 2, 2, 1\}. Initialized from a pretrained dehazing checkpoint, the model is optimized via AdamW (initial learning rate $1 \times 10^{-6}$, weight decay $1 \times 10^{-4}$) using an $\ell_1$ loss and gradient clipping for 20k iterations. The learning rate decays by a factor of 0.5 at 5k, 10k, and 15k iterations. The enhanced images are subsequently processed by the standard 3DGS pipeline with default settings. 

\subsection{MonoSmokeGS}
\label{subsec:monosmokegs}

\begin{center}


\noindent\emph{Linzhe Jiang, Louzhe Xu}


\end{center}
\vspace{-1em}
\begin{figure}[!h]
    \centering
    \includegraphics[width=\linewidth]{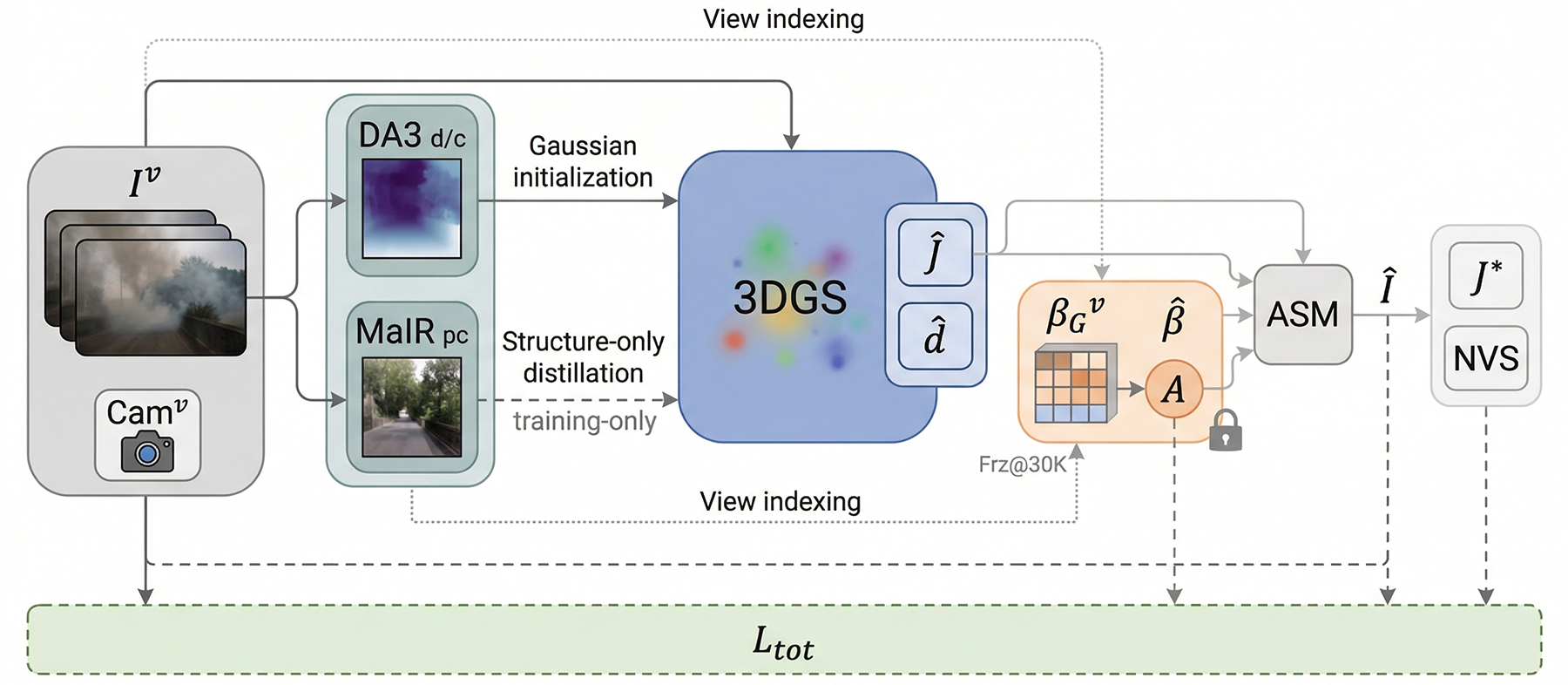}
    \caption{Overview of MonoSmokeGS.}
    \label{fig:monosmokegs}
\end{figure}

\noindent\textbf{Method}~~
They propose MonoSmokeGS, which combines 3DGS with the atmospheric scattering model. Scene Gaussians represent clean appearance and geometry, while a low-resolution beta grid for each view models spatially varying smoke. Depth-AnythingV3 \cite{depthanything3} provides depth and confidence priors for Gaussian initialization and geometry supervision, and MaIR \cite{MaIR} provides structure guidance through gradient-based distillation. The overview of MonoSmokeGS is illustrated in Fig.~\ref{fig:monosmokegs}.

\subsection{CPG-GS: Cascade Prior-Guided Scaffold-GS}
\label{subsec:teamaaa}

\begin{center}


\noindent\emph{Weizhi Nie$^{1}$, Xingan Zhan$^{1}$, Dufeng Zhang$^{1}$}


\noindent\emph{$^{1}$Tianjin University}

\end{center}

\begin{figure}[!hp]
\begin{center}
\includegraphics[width=0.98\linewidth]{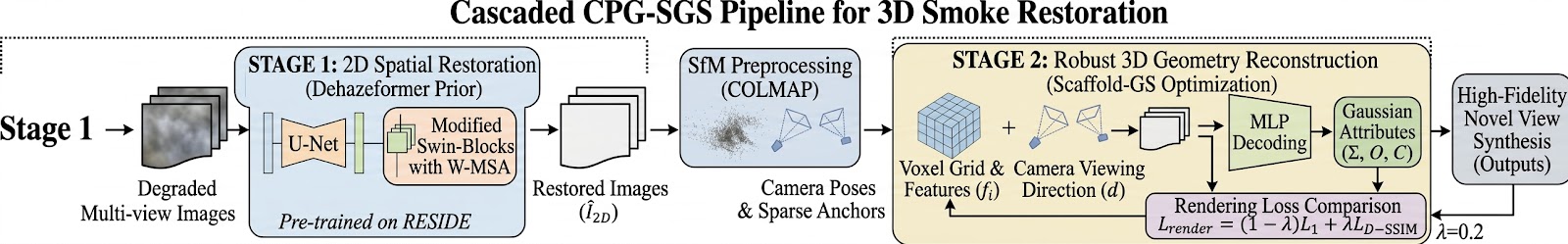} 
\end{center}
\caption{Overall pipeline of CPG-GS. \textbf{Stage 1} utilizes a pre-trained Dehazeformer to remove scattering effects. \textbf{Stage 2} explicitly leverages the restored images (\(\hat{I}_{2D}\)) to extract reliable anchors via COLMAP, followed by per-scene Scaffold-GS optimization.}
\label{fig:AAA}
\end{figure}

\noindent\textbf{Method}~~
The authors propose CPG-GS, a decoupled two-stage cascade framework integrating a 2D spatial prior with a geometry-aware 3D renderer (Fig.~\ref{fig:AAA}). First, a pre-trained Dehazeformer \cite{song2023vision} removes scattering effects from the input, predicting intermediate clean images \(\hat{I}_{2D}\). Then, Scaffold-GS \cite{scaffoldgs} is utilized for robust view synthesis. It initializes sparse 3D anchors from the SfM point cloud, with each anchor storing a learnable neural feature \(f_i\). Given a viewing direction \(d\), a MLP dynamically decodes these features into local 3D Gaussians:
\(
\Sigma, O, C = \text{MLP}(f_i, d).
\)\\


\noindent\textbf{Training Details}~~
The anchor features and MLP are trained with the Adam optimizer for 30,000 iterations, using learning rates of 0.005 and 0.001, respectively. The rendering loss combines \(L_1\) and D-SSIM (\(\lambda = 0.2\)).


\subsection{DH-GS: Depth-Guided Haze-Aware 3DGS for Foggy Scene Reconstruction}
\label{subsec:hnu}

\begin{center}


\noindent\emph{Boyuan Tian$^{1}$, Jingshuo Zeng$^{1}$}


\noindent\emph{$^{1}$Hunan University}

\end{center}


\noindent\textbf{Method}~~
The authors propose a depth-guided 3DGS framework using monocular depth as a structural prior and image dehazing for appearance enhancement. DehazeFormer \cite{song2023vision} generates a dehazed image as the primary 3DGS input, while Depth-AnythingV2 \cite{depth_anything_v2} provides a raw monocular depth map for auxiliary geometric guidance. To preserve relative depth ordering without explicit scale alignment, a Charbonnier loss enforces consistency between per-image min-max normalized rendered and raw depths:
An EMA-based per-Gaussian depth state measures cross-view inconsistency, enabling the conservative pruning of Gaussians that exhibit high inconsistency, low opacity, and large spatial scales. Additionally, a haze-aware regularization term (\(\mathcal{L}_{haze}\)) suppresses erroneous near-field reconstructions in distant regions by penalizing rendered depths that appear systematically closer than expected. 

\subsection{TD-GS: Two-stage Dehazing 3DGS}
\label{subsec:ffffyb}

\begin{center}


\noindent\emph{Yubao Fu$^{1}$ \quad $^{1}$Hainan University}


\end{center}


\noindent\textbf{Method}~~
The authors propose a two-stage pipeline coupling 2D dehazing with per-scene 3DGS. Stage A performs supervised 2D dehazing using a compact U-Net \cite{ronneberger2015u} featuring three pooling stages, double convolution blocks, and a 1×1 convolutional output head. Stage B optimizes a scene-specific set of 3D Gaussians supervised by the dehazed multi-view images. Full-resolution dehazing is emphasized prior to 3D optimization to mitigate detail loss caused by heavy 2D downsampling.\\



\noindent\textbf{Training Details}~~
Images are aspect-ratio resized, center-cropped, and normalized to \([0,1]\). The model is optimized using AdamW with combined L1 and SSIM photometric losses.
The optimization follows the standard 3DGS formulation with adaptive densification and opacity control.


\section{Acknowledgements}
This work was partially supported by the Humboldt Foundation. We sincerely thank the NTIRE 2026 sponsors: OPPO, Kuaishou, and the University of Wurzburg (Computer Vision Lab).

{
    \small
    \bibliographystyle{ieeenat_fullname}
    \bibliography{main}
}

\end{document}